\documentclass{fairmeta}
\usepackage{wrapfig}
\usepackage{placeins}
\usepackage{needspace}

\usepackage{amsmath,amsfonts,bm}

\def\eqref#1{equation~\ref{#1}}

\def\1{\bm{1}}

\DeclareMathAlphabet{\mathsfit}{\encodingdefault}{\sfdefault}{m}{sl}
\SetMathAlphabet{\mathsfit}{bold}{\encodingdefault}{\sfdefault}{bx}{n}

\newcommand{\Plus}{\texttt{+}}

\usepackage{graphicx}
\usepackage{booktabs}
\usepackage{makecell}
\usepackage{subcaption}
\usepackage{hyperref}
\usepackage{url}
\usepackage{array}
\usepackage{xcolor}
\usepackage{tikz}
\usetikzlibrary{calc}
\usepackage[table]{xcolor}
\usepackage{amssymb}
\usepackage{comment}
\usepackage{lipsum}
\usepackage{enumitem}
\usepackage{multirow}
\usepackage{xcolor}
\usepackage{caption}
\usepackage{pifont}
\usepackage[table]{xcolor}
\usepackage{longtable}
\usepackage[T1]{fontenc}
\usepackage{textcomp}
\usepackage[var0]{inconsolata}
\usepackage{inconsolata}
\usepackage{tabularx}
\usepackage{ragged2e}
\usepackage{listings}
\usepackage{xcolor} 
\usepackage{multicol}
\usepackage{adjustbox}
\usepackage[T1]{fontenc}
\usepackage{mdframed}
\usepackage[table]{xcolor}
\definecolor{faintblue}{RGB}{240,245,255}
\usepackage{url}
\usepackage{float}     
\usepackage{tcolorbox}
\usepackage{adjustbox}
\usepackage{pifont}
\usepackage{makecell}

\newcolumntype{Y}{>{\RaggedRight\arraybackslash}X}
\newcommand{\code}[1]{\texttt{\nolinkurl{#1}}}

\definecolor{monokai-bg}{RGB}{255, 254, 250}
\definecolor{monokai-fg}{RGB}{2, 2, 2}
\definecolor{monokai-comment}{RGB}{117,113,94}
\definecolor{monokai-keyword}{RGB}{249,38,114}
\definecolor{monokai-string}{RGB}{230, 169, 116}
\definecolor{monokai-function}{RGB}{73, 179, 114}
\definecolor{monokai-number}{RGB}{174,129,255}
\definecolor{monokai-class}{RGB}{0, 118, 207}
\definecolor{monokai-decorator}{RGB}{253,151,31}

\definecolor{md-bg}{RGB}{255, 254, 250}
\definecolor{md-text}{RGB}{36,41,47}
\definecolor{md-heading}{RGB}{74, 72, 72}
\definecolor{md-code-bg}{RGB}{246,248,250}
\definecolor{md-code}{RGB}{39, 122, 217}

\newcommand{\bench}[1]{\textbf{#1}}
\newcommand{\airsbench}{\textsc{AIRS-Bench}}
\newcommand{\airadojo}{\textsc{AIRA-dojo}}
\newcommand{\mlgym}{\textsc{MLGym}}
\newcommand{\opdraft}{\textsc{Draft}}
\newcommand{\opdebug}{\textsc{Debug}}
\newcommand{\opimprove}{\textsc{Improve}}
\newcommand{\opanalyze}{\textsc{Analyze}}

\title{\airsbench: a Suite of Tasks for Frontier AI Research Science Agents}

\author[1,2,*]{Alisia Lupidi}
\author[1,*]{Bhavul Gauri}
\author[1,2,*]{Thomas Simon Foster}
\author[1,*]{Bassel Al Omari}
\author[1,*]{Despoina Magka}
\author[1]{Alberto Pepe}
\author[1]{Alexis Audran-Reiss}
\author[1,\dagger]{Muna Aghamelu}
\author[1]{Nicolas Baldwin}
\author[1]{Lucia Cipolina-Kun}
\author[1]{Jean-Christophe Gagnon-Audet}
\author[1]{Chee Hau Leow}
\author[1]{Sandra Lefdal}
\author[1]{Hossam Mossalam}
\author[1,\dagger]{Abhinav Moudgil}
\author[1]{Saba Nazir}
\author[1,\dagger]{Emanuel Tewolde}
\author[1]{Isabel Urrego}
\author[1]{Jordi Armengol Estape}
\author[1]{Amar Budhiraja}
\author[1]{Gaurav Chaurasia}
\author[1]{Abhishek Charnalia}
\author[1]{Derek Dunfield}
\author[1,3]{Karen Hambardzumyan}
\author[1]{Daniel Izcovich}
\author[1]{Martin Josifoski}
\author[1]{Ishita Mediratta}
\author[1]{Kelvin Niu}
\author[1]{Parth Pathak}
\author[1]{Michael Shvartsman}
\author[1,3]{Edan Toledo}
\author[1]{Anton Protopopov}
\author[1,\dagger]{Roberta Raileanu}
\author[1]{Alexander Miller}
\author[1]{Tatiana Shavrina}
\author[1,2]{Jakob Foerster}
\author[1]{Yoram Bachrach}

\affiliation[1]{FAIR at Meta}
\affiliation[2]{University of Oxford}
\affiliation[3]{University College London}

\contribution[*]{Joint first author}
\contribution[\dagger]{Work done at Meta}

\abstract{

LLM agents hold significant promise for advancing scientific research.
To accelerate this progress, we introduce \airsbench~(the \textit{AI Research Science Benchmark}), a suite of 20 tasks sourced from state-of-the-art machine learning papers.
These tasks span diverse domains, including language modeling, mathematics, bioinformatics, and time series forecasting.
\airsbench~tasks assess agentic capabilities over the full research lifecycle---including idea generation, experiment analysis and iterative refinement---without providing baseline code.
The \airsbench~task format is versatile, enabling easy integration of new tasks and rigorous comparison across different agentic frameworks.
We establish baselines using frontier models paired with both sequential and parallel scaffolds.
Our results show that agents exceed human SOTA in four tasks but fail to match it in sixteen others.
Even when agents surpass human benchmarks, they do not reach the theoretical performance ceiling for the underlying tasks.
These findings indicate that \airsbench~is far from saturated and offers substantial room for improvement.
We open-source the \airsbench~task definitions and evaluation code to catalyze further development in autonomous scientific research.
}

\date{\today}
\correspondence{Despoina Magka at \email{despoinam@meta.com}, Yoram Bachrach at \email{yorambac@meta.com}, Jakob Foerster at \email{jnf@meta.com}}

\metadata[Code]{\url{https://github.com/facebookresearch/airs-bench}}

\begin{document}

\maketitle

\section{Introduction}
From the rise of deep learning through to the current era of Large Language models (LLMs), the rapid progress in machine learning (ML) has been largely benchmark-driven, making robust evaluations a key requirement~\citep{hardt2025emerging}.
This trend is reinforced by the current \emph{reviewing crisis}, which highlights the limits of human judgment in assessing the quality of research contributions  ~\citep{Xu2022, lawrence2022}.
Recent improvements in LLM technology have extended their abilities from simple tasks to complex agentic workflows, including scientific reasoning and coding~\citep{khatri2025artscalingreinforcementlearning, andrews2025arescalingagentenvironments, lu2024aiscientistfullyautomated, yamada2025aiscientistv2workshoplevelautomated}.
Whilst early LLM research focused on improving the capabilities of LLMs through better data and pre-training, recent work has focused on extending their abilities by leveraging more test-time compute~\citep{NEURIPS2023_271db992}, in particular by leveraging \emph{scaffolds} that iteratively query the LLM at test-time so as to obtain a better response. Such scaffolds can enable a test-time search of the solution space by incorporating environment feedback~\citep{nathani2025mlgym, toledo2025airesearchagentsmachine} and yield much more performant solutions. Despite this potential, we lack a standardized framework to measure how well these agents perform the actual work of a research scientist.  To fill this gap, we introduce \airsbench.



We adopt the broadly accepted definition of an agent as a \textit{computer system situated in an environment that is capable of autonomous action in order to meet its design objectives} \citep{WoJe95}. In our context, an \textit{agent} is characterized as an \textit{LLM} augmented by a \textit{scaffold}. The LLM is the underlying probabilistic model that serves as the agent's primary reasoning core, either self-hosted OSS model or API-based. The scaffold acts as an orchestrating layer, providing the necessary mechanisms to translate LLM outputs into systematic exploration of the solution space. The scaffold is typically implemented by a \textit{harness}, the execution framework responsible for instantiating and managing diverse scaffolding configurations. Examples are the ReAct scaffold~\citep{yao2023react}, implemented within the \mlgym~harness~\citep{nathani2025mlgym} or the Monte Carlo Tree Search (MCTS) scaffold~\citep{Kocsis2006}, implemented within the \airadojo~harness~\citep{toledo2025airesearchagentsmachine}. We experiment with both sequential (linear) and parallel scaffolds. Sequential scaffolds follow a linear execution loop, where each LLM query is conditioned on feedback from previous actions~\citep{yao2023react, nathani2025mlgym}. Parallel scaffolds, by contrast, maintain and grow a population of potential solutions, utilizing data structures such as trees to guide exploration~\citep{novikov2025alphaevolvecodingagentscientific,openevolve,toledo2025airesearchagentsmachine}.

A growing area of research for LLM agents is represented by the automation of AI research itself. We call agents designed to automate and accelerate AI research \textit{AI Research Agents}. 
Accurately evaluating the performance of such agents, however, remains a significant challenge. ML has long struggled with reproducibility and statistical noise \citep{recht2018cifar10classifiersgeneralizecifar10, hardt2025emerging}, and the agentic paradigm adds further complications:

\begin{itemize}
    \item Data contamination: LLMs are trained on vast amounts of internet data, and they often memorize benchmark solutions. This makes it difficult to assess whether an agent is truly ``reasoning'' on a task.

    \item Environmental standardization: agentic environments are difficult to standardize across different studies, and it is often unclear if success comes from the agent's capabilities or the specific way the environment was built.

    \item  Computational cost: The high computational overhead of each autonomous run makes it difficult to conduct the extensive trials necessary to obtain statistically significant results.

\end{itemize}

These factors have exacerbated the \textit{evaluation crisis} of AI Research Agents, as performance on existing benchmarks is increasingly obscured by pretraining leakage, inconsistent environment setups, and noisy empirical evaluations \citep{haimes2024benchmarkinflationrevealingllm, dehghani2021benchmarklottery}.
\begin{figure}
    \centering
    \includegraphics[width=\linewidth]{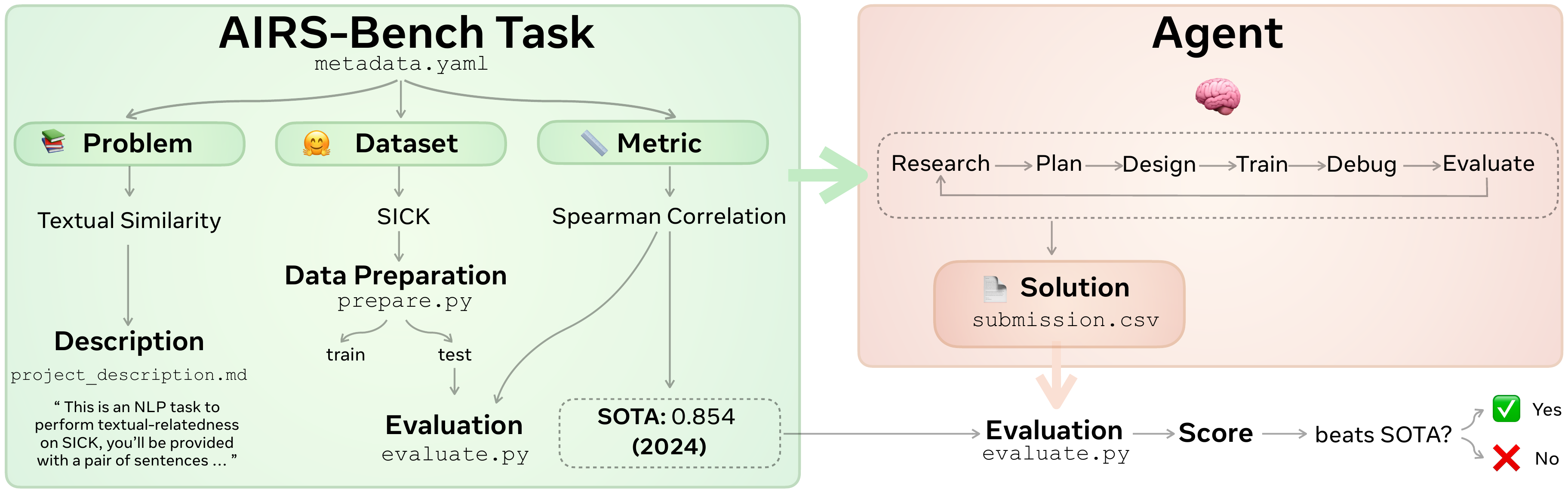}
    \caption{Example of an \airsbench{} task. Each task  is specified by a \{problem, dataset, metric\} triplet. The \textbf{problem} defines the core computational challenge to be solved (e.g. textual similarity); the \textbf{dataset} specifies which data to solve the challenge over (e.g. SICK); finally, the \textbf{metric} is used to quantify performance (e.g. Spearman correlation). The agent receives the full task specification and is expected to develop a solution that in most cases generates predictions on the test labels file, which are then evaluated and compared with the state-of-the-art result.
    }
    \label{fig:whatisatask}
\end{figure}
We introduce the \textit{AI Research Science Benchmark} (\airsbench) to address these limitations and provide a standardized evaluation of AI Research Agents. The philosophy behind \airsbench~is that, much like the progress in ML, the advancement of AI Research Agents should be a benchmark-driven process. \airsbench~ is a suite of 20 agentic tasks, curated from recent state-of-the-art (SOTA) literature. This ensures that tasks are both challenging and relevant to the broader ML community. In our design, agents are not required to tackle these tasks through direct inference. Instead, they must generate the code necessary to train and validate an ML model. We then evaluate the agent's capabilities by executing the code it generated and measuring the model performance. This approach effectively evaluates the agent's capacity to function as an autonomous research scientist.

To both standardize evaluations and make the tasks closely match their original versions in the ML literature, we developed a \textit{task configuration standard} that encapsulates an \airsbench{} task (see Figure~\ref{fig:whatisatask} for an example) and a task creation pipeline based on semi-manual sourcing, creation, reviewing and verification of tasks. Our proposed \airsbench~task standard can be adapted and extended to virtually any ML problem, effectively democratizing AI agentic research.
For fair evaluations across all agents we have designed our experiments to make runs as comparable as possible by factoring in infrastructure issues (e.g. crashing runs) and eliminating environment discrepancies. To investigate the impact of different scaffolds, we evaluate both the~\airadojo ~\citep{toledo2025airesearchagentsmachine} and~\mlgym ~\citep{nathani2025mlgym} harnesses combined with different LLMs (CWM~\citep{carbonneaux2025cwm}, GPT-4o~\citep{openai2024gpt4osystemcard}, gpt-oss-20b and gpt-oss-120b~\citep{openai2024gptoss}, o3-mini~\citep{openai2024o3o4mini}, Devstral~\citep{rastogi2025devstral}).
We also introduce an evaluation protocol, consisting of different metrics and aggregating results across seeds and tasks to allow for statistically robust and interpretable results.


The key contributions of the paper are:
\begin{itemize}
    \item \textbf{Uncontaminated benchmark for AI Research Agents}: \airsbench~uniquely assesses agents on the end-to-end ML research workflow: idea generation, methodology design, experiment analysis, and iterative refinement---without access to baseline code. This enables a realistic evaluation of agentic research abilities. \airsbench~tasks cover a variety of machine learning research problems, ranging from NLP, math and code to biochemical modeling and time series forecasting. The benchmark is designed to be compatible with multiple agentic frameworks (i.e. harnesses), supporting robust and fair comparisons across different agent architectures and tool integrations.
    \item \textbf{Task configuration standard}: we introduce a task configuration standard and fixed evaluation metrics, to ensure reproducibility and minimize runtime and environment inconsistencies. We establish quality by leveraging human checks throughout task building, reviewing, and verification.
    \item \textbf{Empirical analysis}: we benchmark all tasks across frontier open and closed-source models  and different scaffolds, revealing substantial variation in performance. We closely inspect cases where agents are shown to match or beat human SOTA and analyze the agent-generated solutions---including cases where agents discover effective SOTA-exceeding combinations of approaches.
    \item \textbf{Evaluation protocol}: we define a suite of metrics assessing different aspects of agents' capabilities including valid submission rates, normalized performance scores, and Elo ratings. Normalization enables us to aggregate performance across tasks by applying transforms that map $0.0$ scores to the weakest valid solution and $1.0$ to human SOTA.  The metrics used are designed to accurately reflect progress on developing state-of-the-art AI Research Agents. We report results on these metrics across the range of selected agents; our results suggest that the benchmark is far from solved, leaving plenty of headroom for development of AI Research Agents.
\end{itemize}



The rest of the paper is organized as follows. Section~\ref{sec:related_work} contains an overview of existing benchmarks within the AI research domain. Section~\ref{sec:agents} introduces the agent, scaffold and harness definitions and section~\ref{sec:method} presents the benchmark structure and creation methodology. Sections~\ref{sec:exp_design} and~\ref{sec:exp_results} outline our experimental design and results. We conclude with discussing  learnings and key findings in Section~\ref{sec:discussion}.

\section{Related Work}
\label{sec:related_work}

We examined a number of benchmarks evaluating the ability of AI agents to carry out scientific research, including ML-related agentic tasks. Below, we outline several trends we identified and how they relate with \airsbench.

\textbf{Full cycle of Scientific Method}. Many earlier papers create agentic tasks from available sources: Github repositories (CSR-bench;~\citealp{xiao2025csrbenchbenchmarkingllmagents}), top-tier conference papers and their data (PaperBench;~\citealp{starace2025paperbenchevaluatingaisability}), Kaggle competitions (MLE-bench;~\citealp{chan2025mlebenchevaluatingmachinelearning}, DSBench;~\citealp{jing2025dsbenchfardatascience}), ML research competitions (MLRC-Bench;~\citealp{zhang2025mlrcbenchlanguageagentssolve}), and carefully selected cross-domain papers (SciReplicate-bench;~\citealp{xiang2025scireplicatebenchbenchmarkingllmsagentdriven}). The experimental cycle of these benchmarks can include ideation, implementation, experimentation, analysis and comparison to previous results, with several benchmarks focusing on separate stages of the cycle, such as IdeaBench \citep{guo2024ideabenchbenchmarkinglargelanguage}, LiveIdeaBench \citep{ruan2025liveideabenchevaluatingllmsdivergent}, AI Idea Bench \citep{qiu2025aiideabench2025} and ResearchBench \citep{liu2025researchbenchbenchmarkingllmsscientific} for ideation, FML-bench~\citep{zou2025fmlbenchbenchmarkautomaticml} for idea novelty estimation, SurveyBench \citep{sun2025surveybenchllmagentswriteacademic} for literature review, and DataGovBench \citep{liu2025datagovbenchbenchmarkingllmagents}, DCA-Benchmark \citep{huang2025dcabenchbenchmarkdatasetcuration}, DA-Code \citep{huang2024dacodeagentdatascience}, and DS-1000 \citep{lai2022ds1000naturalreliablebenchmark} for data quality, implementation and analysis. \airsbench, on the other hand, requires the agent to excel in every step of the scientific method to perform well on the benchmark tasks, spanning hypothesis generation, implementation, experimentation, and analysis.


\begin{table}[t]

\centering

\resizebox{\textwidth}{!}{%

\renewcommand{\arraystretch}{2.0}

\begin{tabular}{
    c>{\raggedright}
    c
    c
    c
    @{\hspace{2cm}}
    p{0.7cm}
    p{0.7cm}
    p{0.7cm}
    p{0.7cm}
    c
    c
    >{\centering\arraybackslash}p{10cm}
}

\toprule

\LARGE\textbf{Benchmark} & \LARGE\textbf{Task Composition and} & \LARGE\textbf{AI Research} & \LARGE\textbf{Task} & \multicolumn{4}{c}{\LARGE\textbf{Scientific Method}} & \LARGE\textbf{No} & \LARGE\textbf{Agent} \\ 

\cline{5-8}

& \LARGE\textbf{Origin} & \LARGE\textbf{Data}& \LARGE\textbf{Horizon} & \LARGE\textbf{(H)} & \LARGE\textbf{(I)} & \LARGE\textbf{(E)} & \LARGE\textbf{(A)} & \LARGE\textbf{Baseline} & \LARGE\textbf{Compute} \\ 

\toprule

\LARGE\textbf{AIRS-Bench (ours)} & \LARGE 20 tasks from 17 machine learning & & & & & & & &  \\ 

 & \LARGE papers with state-of-the-art results & \textcolor{green!50!black}{\LARGE\checkmark} & \cellcolor{green!12}\LARGE long & \textcolor{green!50!black}{\LARGE\checkmark} & \textcolor{green!50!black}{\LARGE\checkmark} & \textcolor{green!50!black}{\LARGE\checkmark} & \textcolor{green!50!black}{\LARGE\checkmark} & \textcolor{green!50!black}{\LARGE\checkmark} & \cellcolor{green!12}\LARGE high GPU \\ 

\midrule

\LARGE Automated LLM  & \LARGE 76 tasks from the NanoGPT  & & & & & & & & \\ 

\LARGE  Speedrun~\citep{zhao2025automatedllmspeedrunningbenchmark} & \LARGE Speedrun Github repo & \textcolor{green!50!black}{\LARGE\checkmark} & \cellcolor{green!12}\LARGE long & \textcolor{red!80!black}{\LARGE\textbf{x}} & \textcolor{green!50!black}{\LARGE\checkmark} & \textcolor{green!50!black}{\LARGE\checkmark} & \textcolor{green!50!black}{\LARGE\checkmark} & \textcolor{green!50!black}{\LARGE\checkmark} & \cellcolor{green!12}\LARGE high GPU \\ 

\midrule
\LARGE Auto-Bench & \LARGE 6 graph discovery tasks     & & & & & & & &  \\

\LARGE ~\citep{chen2025autobenchautomatedbenchmarkscientific} & \LARGE from chemistry and social networks   & \textcolor{red!80!black}{\LARGE\textbf{x}} & \cellcolor{red!10}\LARGE short & \textcolor{green!50!black}{\LARGE\checkmark} & \textcolor{red!80!black}{\LARGE\textbf{x}} & \textcolor{red!80!black}{\LARGE\textbf{x}} & \textcolor{red!80!black}{\LARGE\textbf{x}} & \textcolor{green!50!black}{\LARGE\checkmark} & \LARGE not specified \\ 

\midrule
\LARGE CORE-Bench & \LARGE 270 tasks from 90 social, medical, & & & & & & & &  \\

\LARGE ~\citep{siegel2024core} & \LARGE and computer science papers & \textcolor{green!50!black}{\LARGE\checkmark} & \cellcolor{yellow!12}\LARGE medium & \textcolor{red!80!black}{\LARGE\textbf{x}} & \textcolor{green!50!black}{\LARGE\checkmark} & \textcolor{green!50!black}{\LARGE\checkmark} & \textcolor{green!50!black}{\LARGE\checkmark} & \textcolor{red!80!black}{\LARGE\textbf{x}} & \cellcolor{yellow!12}\LARGE low GPU \\ 

\midrule

\LARGE CSR-Bench & \LARGE 107 Github repos from CV, NLP  & & & & & & & &  \\

\LARGE ~\citep{xiao2025csrbenchbenchmarkingllmagents} & \LARGE  and interdisciplinary ML papers & \textcolor{green!50!black}{\LARGE\checkmark} & \LARGE not specified & \textcolor{red!80!black}{\LARGE\textbf{x}} & \textcolor{green!50!black}{\LARGE\checkmark} & \textcolor{green!50!black}{\LARGE\checkmark} & \textcolor{green!50!black}{\LARGE\checkmark} & \textcolor{red!80!black}{\LARGE\textbf{x}}  & \LARGE not specified \\ 

\midrule
\LARGE MLE-Bench & \LARGE 75 Kaggle   & & & & & & & &  \\

\LARGE ~\citep{chanMLEbenchEvaluatingMachine2024} & \LARGE competitions
 & \textcolor{red!80!black}{\LARGE\textbf{x}} & \cellcolor{green!12}\LARGE long & \textcolor{green!50!black}{\LARGE\checkmark} & \textcolor{green!50!black}{\LARGE\checkmark} & \textcolor{green!50!black}{\LARGE\checkmark} & \textcolor{green!50!black}{\LARGE\checkmark} & \textcolor{green!50!black}{\LARGE\checkmark} & \cellcolor{green!12}\LARGE high GPU \\ 

\midrule
 \LARGE MLGym-Bench & \LARGE 13 tasks from supervised learning,  & & & & & & & &  \\

  \LARGE ~\citep{nathani2025mlgym} & \LARGE RL and algorithmic reasoning problems & \textcolor{green!50!black}{\LARGE\checkmark} & \cellcolor{yellow!12}\LARGE medium & \textcolor{green!50!black}{\LARGE\checkmark} & \textcolor{green!50!black}{\LARGE\checkmark} & \textcolor{green!50!black}{\LARGE\checkmark} & \textcolor{green!50!black}{\LARGE\checkmark} & \textcolor{green!50!black}{\LARGE\checkmark} & \cellcolor{green!12}\LARGE high GPU \\ 

  \midrule

\LARGE ML-Agent-Bench &  \LARGE 13 tasks from Kaggle, computer & & & & & & & &  \\

\LARGE ~\citep{tangMLBenchEvaluatingLarge2024} &  \LARGE science papers, and textbooks & \textcolor{green!50!black}{\LARGE\checkmark} & \cellcolor{yellow!12}\LARGE medium & \textcolor{green!50!black}{\LARGE\checkmark} & \textcolor{green!50!black}{\LARGE\checkmark} & \textcolor{green!50!black}{\LARGE\checkmark} & \textcolor{green!50!black}{\LARGE\checkmark} & \textcolor{red!80!black}{\LARGE\textbf{x}} & \cellcolor{yellow!12}\LARGE low GPU \\ 

  \midrule

\LARGE SWE-Bench & \LARGE 2,294 Github issues from & & & & & & & &  \\

\LARGE ~\citep{jimenez2023swe} & \LARGE  open-source repos & \textcolor{red!80!black}{\LARGE\textbf{x}} & \cellcolor{red!10}\LARGE short & \textcolor{red!80!black}{\LARGE\textbf{x}} & \textcolor{green!50!black}{\LARGE\checkmark} & \textcolor{green!50!black}{\LARGE\checkmark} & \textcolor{green!50!black}{\LARGE\checkmark} & \textcolor{green!50!black}{\LARGE\checkmark} & \cellcolor{red!10}\LARGE CPU \\ 

\midrule

 \LARGE SciReplicate-Bench & \LARGE 100 tasks from   & & & & & & & &  \\

  \LARGE ~\citep{Xiang2025SciReplicateBenchBL} & \LARGE  36 NLP papers  & \textcolor{green!50!black}{\LARGE\checkmark} & \cellcolor{yellow!12}\LARGE medium & \textcolor{red!80!black}{\LARGE\textbf{x}} & \textcolor{green!50!black}{\LARGE\checkmark} & \textcolor{green!50!black}{\LARGE\checkmark} & \textcolor{green!50!black}{\LARGE\checkmark} & \textcolor{red!80!black}{\LARGE\textbf{x}} & \LARGE not specified \\ 

\midrule

\LARGE PaperBench & \LARGE 8,316 rubrics from
& & & & & & & &  \\

\LARGE ~\citep{starace2025paperbench} & \LARGE  20 ICML papers
& \textcolor{green!50!black}{\LARGE\checkmark} & \cellcolor{red!10}\LARGE short & \textcolor{red!80!black}{\LARGE\textbf{x}} & \textcolor{green!50!black}{\LARGE\checkmark} & \textcolor{green!50!black}{\LARGE\checkmark} & \textcolor{green!50!black}{\LARGE\checkmark} & \textcolor{red!80!black}{\LARGE\textbf{x}} & \cellcolor{yellow!12}\LARGE low GPU \\ 

\midrule

\LARGE ResearchBench & \LARGE Hypothesis generation task & & & & & & & &  \\

\LARGE ~\citep{liu2025researchbench} & \LARGE for 1386 papers across the sciences & \textcolor{green!50!black}{\LARGE\checkmark} & \LARGE not specified & \textcolor{green!50!black}{\LARGE\checkmark} & \textcolor{red!80!black}{\LARGE\textbf{x}} & \textcolor{red!80!black}{\LARGE\textbf{x}} & \textcolor{green!50!black}{\LARGE\checkmark} & \textcolor{green!50!black}{\LARGE\checkmark} & \LARGE not specified \\ 

\midrule

\LARGE RE-Bench & \LARGE 7 LLM pretraining/coding tasks  & & & & & & & &  \\

\LARGE ~\citep{rebench-metr} & \LARGE and 71 human attempts & \textcolor{red!80!black}{\LARGE\textbf{x}} & \cellcolor{yellow!12}\LARGE medium & \textcolor{green!50!black}{\LARGE\checkmark} & \textcolor{green!50!black}{\LARGE\checkmark} & \textcolor{green!50!black}{\LARGE\checkmark} & \textcolor{green!50!black}{\LARGE\checkmark} & \textcolor{red!80!black}{\LARGE\textbf{x}} & \cellcolor{green!12}\LARGE high GPU\\ 

\midrule

\LARGE LMR-Bench  & \LARGE 28 code reproduction tasks  & & & & & & & &  \\

\LARGE ~\citep{yan2025lmrbenchevaluatingllmagents} & \LARGE from 23 NLP papers & \textcolor{green!50!black}{\LARGE\checkmark} & \LARGE not specified & \textcolor{red!80!black}{\LARGE\textbf{x}}  & \textcolor{green!50!black}{\LARGE\checkmark} & \textcolor{red!80!black}{\LARGE\textbf{x}}  & \textcolor{red!80!black}{\LARGE\textbf{x}}  & \textcolor{red!80!black}{\LARGE\textbf{x}} & \LARGE not specified \\ 

\midrule

\LARGE PostTrainBench & \LARGE Post-train 4 base LLMs to  & & & & & & & &  \\

\LARGE ~\citep{posttrainbench_2025} & \LARGE maximise perf across 5 benchmarks & \textcolor{red!80!black}{\LARGE\textbf{x}} & \cellcolor{yellow!12}\LARGE medium & \textcolor{green!50!black}{\LARGE\checkmark} & \textcolor{green!50!black}{\LARGE\checkmark} & \textcolor{green!50!black}{\LARGE\checkmark} & \textcolor{green!50!black}{\LARGE\checkmark} & \textcolor{red!80!black}{\LARGE\textbf{x}} & \cellcolor{green!12}\LARGE high GPU\\ 

\bottomrule
\end{tabular}
}

\caption{Comparison of 14 popular agentic AI research benchmarks with \airsbench~across key evaluation dimensions. Task horizon refers to the time window provided to solve the problem and it can be short (<1 hour), medium (1-12 hours) or long (>12 hours). We indicate whether the benchmarks assess for the hypothesis generation (\textbf{H}), implementation (\textbf{I}), experimentation (\textbf{E}) and analysis (\textbf{A}) stages of the scientific pipeline. Compute refers to the resources provided to the agent and it can be CPU, low GPU ($\leq$ 1 hour per task) and high GPU (>1 hour per task). No baseline refers to whether the agent has access to a baseline solution to tackle the problem.}

\label{tab:benchmarks}

\end{table}

\textbf{Access to Baseline Solution}. Benchmarks differ significantly in two ways: whether the agent is granted access to a baseline solution, and the degree of ``saturation'' within the underlying problem. \airsbench~prioritizes unsaturated tasks, and does not provide a starter solution to the agent. This approach makes \airsbench~tasks challenging, as agents must navigate a longer reasoning horizon to make progress independently.

\textbf{Different environments}. Recent benchmarks leverage a wide range of different environments. While most setups closely resemble \airsbench---requiring agents to interact via prompts and code execution---others incorporate gamified environments (DiscoveryWorld;~\citealp{jansen2024discoveryworldvirtualenvironmentdeveloping}) or physics engines (FEABench;~\citealp{mudur2024feabench}). Task sets are typically manually curated, though some benchmarks (DiscoveryBench;~\citealp{majumder2024discoverybenchdatadrivendiscoverylarge}, SUPER;~\citealp{bogin2024superevaluatingagentssetting}) also include synthetic tasks. Some papers simulate user interactions within the environment (AppWorld;~\citealp{trivedi2024appworldcontrollableworldapps}).

\textbf{Diversity of domains}. Beyond core ML tasks, there is a growing trend to expand agentic evaluation into broader scientific domains, such as bioinformatics, chemistry, and physics. We identify more than 20 different scientific domains covered across recent benchmarks, including DiscoveryWorld~\citep{jansen2024discoveryworldvirtualenvironmentdeveloping}, DiscoveryBench~\citep{majumder2024discoverybenchdatadrivendiscoverylarge}, CURIE~\citep{cui2025curieevaluatingllmsmultitask}, FEABench~\citep{mudur2024feabench}, ScienceAgentBench~\citep{chen2025scienceagentbenchrigorousassessmentlanguage}, CORE-bench~\citep{siegel2024corebenchfosteringcredibilitypublished}, AUTObench~\citep{chen2025autobenchautomatedbenchmarkscientific},  ResearchBench~\citep{liu2025researchbenchbenchmarkingllmsscientific} and BioMLbench~\citep{biomlbench2025}. \airsbench~ features text and tabular-based tasks across seven categories of ML problems, including language modeling, mathematics, code generation, molecular modeling, and time-series forecasting. Currently, \airsbench~provides a comprehensive testbed for agents, with the potential for future expansion into additional scientific domains.

We selected a subset of benchmarks that most closely resemble \airsbench—specifically those evaluating an agent's ability to conduct AI research independently. Table~\ref{tab:benchmarks} summarizes this comparison across several dimensions: task composition and origin, reasoning horizon, the stages of the scientific method covered, access to baseline solutions, and compute requirements. \airsbench{} is composed of tasks grounded in AI research data and crafted so that AI research agents (i) tackle long-horizon research challenges that mirror the complexity of scientific discovery (ii)  engage with the complete scientific pipeline from hypothesis generation to implementation, experimentation and analysis in an iterative fashion (iii) operate without a starting solution to ensure unbiased evaluation and promote original research and (iv) leverage substantial computational resources to enable thorough solution space exploration and tackle modern AI research problems. These characteristics make \airsbench~well-suited for tracking progress toward developing state-of-the-art AI Research Agents.




\section{Agents, Scaffolds, Harnesses}
\label{sec:agents}

\begin{figure}[htbp]
    \centering
    \includegraphics[width=\linewidth]{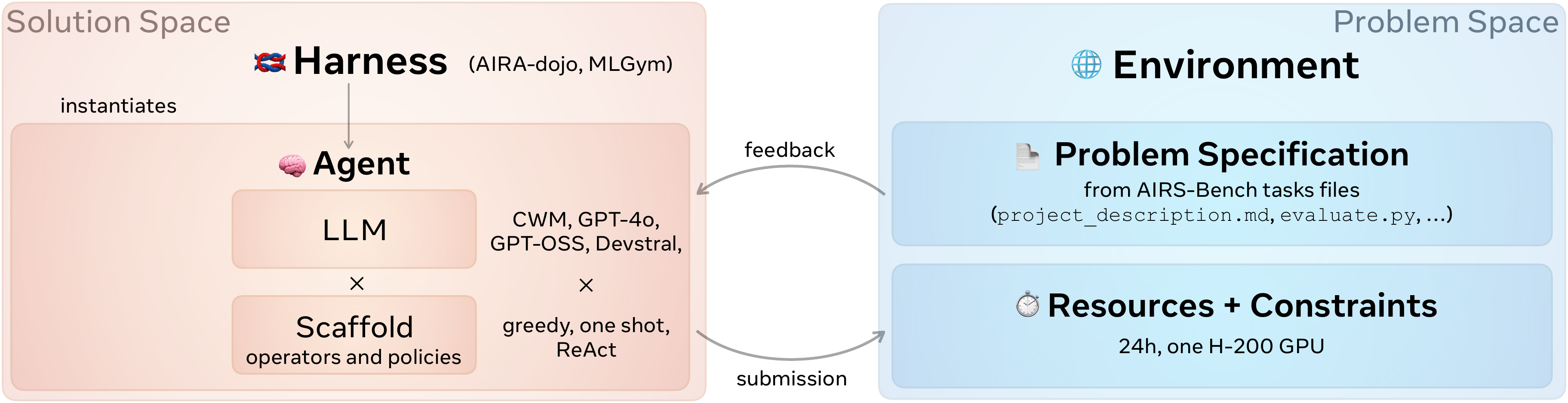}
    \caption{We define an \textbf{agent} as a pair consisting of a large language model (LLM) and a scaffold. A \textbf{scaffold} comprises a set of mechanisms, such as operators and search algorithms, that enable the LLM to explore the solution space effectively. Scaffolds are instantiated by a \textbf{harness}, which serves as a system that encapsulates the agent and manages its research process. The \textbf{environment} provides the agent with the problem specifications, as well as any constraints and resources available for its exploration.}
    \label{fig:agentscaffoldharness}
\end{figure}




In line with the broader agent research literature, we view an agent as a computer system that is situated in some environment and and is able to act autonomously in this environment in order to achieve its design objectives \citep{WoJe95}. For our specific context of LLM agents for AI Research, we define an \textit{agent} as the combination of an LLM and a  \textit{scaffold}. A \textit{scaffold} is an algorithm, expressed as a a set of operators and search policies governing how the agent searches the space of possible solutions for the task at hand. Scaffolds are instantiated within a \textit{harness}—a system that wraps the agent and manages its execution (see Figure~\ref{fig:agentscaffoldharness}). Examples of harnesses include AIDE~\citep{weco2024aide, jiang2025aideaidrivenexplorationspace}, Claude Code,~\footnote{\url{https://code.claude.com/docs/}} OpenCode,~\footnote{\url{https://github.com/anomalyco/opencode}} \mlgym~\citep{nathani2025mlgym} and \airadojo~\citep{toledo2025airesearchagentsmachine}. For instance, we refer to greedy search as a scaffold that is instantiated within the \airadojo~harness.

\textbf{\airadojo~}is a harness that enables the agent to evolve its solution through a set of operators and a search policy. The search policy (e.g., greedy search, Monte Carlo Tree Search~\citep{Kocsis2006}, evolutionary algorithms~\citep{novikov2025alphaevolvecodingagentscientific,openevolve,FunSearch2023}) guides the exploration of the solution space, while the operators modify existing solutions to generate new candidate solutions. This process results in a tree structure, where each node represents a Python code solution, created by one of the
following operators: (1) \textit{Draft}, which generates the initial set of solutions; (2) \textit{Debug}, which identifies and corrects errors within a given node; and (3) \textit{Improve}, which enhances a solution to increase its performance according to specified evaluation criteria. \airadojo~operators enhance AIDE to (i) promote solution diversity, by inserting into the context solutions of sibling nodes, and (ii) facilitate bug fixing, by including in the context the entire ancestral memory of the solution's debug chain. \airadojo~allows access to the internet, including additional data and pretrained models, which are loaded in the cache for the agent to use according to prompting instructions. \airadojo's operator prompts can be found in Appendix~\ref{app:airadojo_details}.

\textbf{\mlgym~}is a harness through which an agent can sequentially improve its initial solution in a ReAct-like manner \citep{yao2023react}. The agent can develop a solution based on its own ideation and feedback from the execution of the implementation. Access to the internet is also allowed in a manner similar to \airadojo, with the option to cache pretrained models and instruct the agent about them in the prompt. The agent has access to tools and bash, and the full system prompt is present in Appendix~\ref{app:mlgym_details}.

\section{Method}
\label{sec:method}

\bench{\airsbench} consists of $20$ tasks extracted from $17$ machine learning papers and $16$ different datasets. To ensure selection of highly impactful tasks for the community, we sourced tasks and SOTA results from papers published at well-known AI conferences and journals and arXiv preprints
(see Appendix~\ref{app:task_paper_info} for a distribution of publication venues and years). The initial list of tasks was sourced from PapersWithCode's leaderboards ~\citep{PapersWithCode, kardas2020axcellautomaticextractionresults} and filtered down for datasets that had a state-of-the-art result published between 2020 and 2025; additional criteria applied at this stage included availability of the dataset (preferably via HuggingFace) and existence of a train and a test split. After verifying these requirements, we manually created each task and reviewed the SOTA result associated with it, by confirming the number reported in the cited paper, checking that the evaluation metric and dataset splits were identical and updating the SOTA result if a better score could be found. By following this methodology, we initially created and evaluated approximately 100 tasks from $\sim$85 different machine learning papers and datasets. We subsequently selected 20 out of these to ensure tractability and accuracy of the benchmark. The process is outlined in Appendix~\ref{sec:selection-process}.

\begin{figure}
    \centering
    \includegraphics[width=\linewidth]{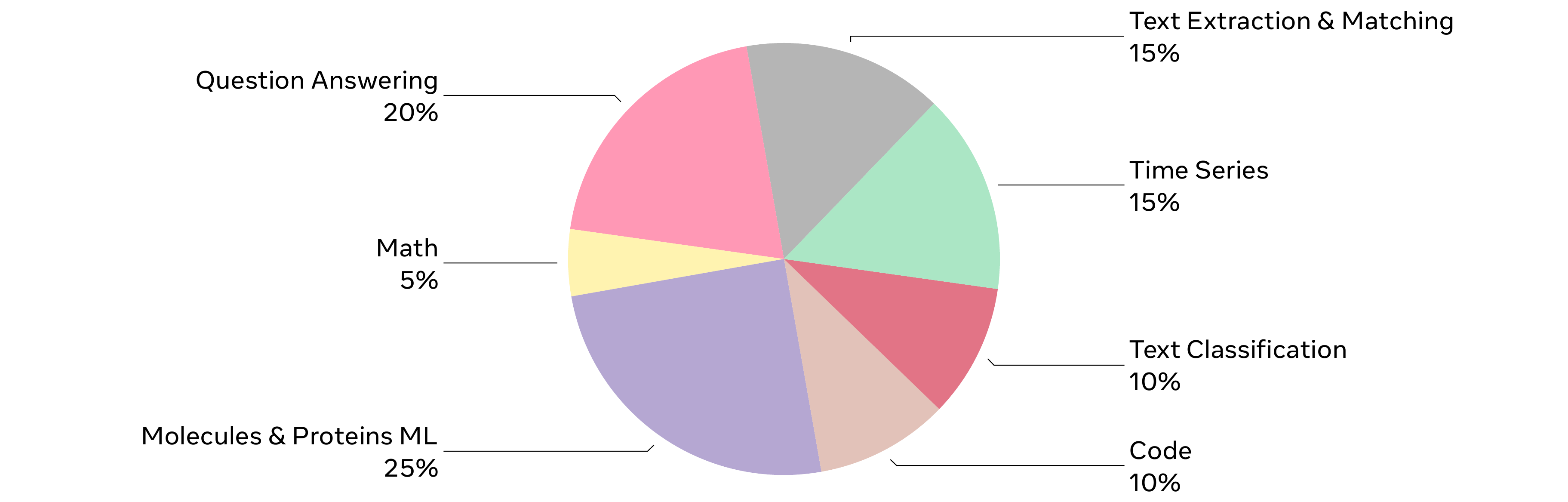}
    \caption{Distribution of \airsbench~tasks by category. We consider 7 distinct task categories in total: \textit{Code}, \textit{Math}, \textit{Molecules \& Proteins ML}, \textit{Question Answering}, \textit{Text Classification}, \textit{Text Extraction \& Matching}, and \textit{Time Series}.
}
    \label{fig:airs_categories_piechart}
\end{figure}

\subsection{Tasks Diversity}

Figure~\ref{fig:airs_categories_piechart} show the $7$ distinct categories that \airsbench{}'s $20$ tasks are organized into, with the tasks spanning a broad spectrum of ML problems. The largest share is dedicated to the NLP domain, comprising \textit{Question Answering} (4/20, open-ended and extractive question answering over diverse contexts), \textit{Text Extraction \& Matching} (3/20, e.g. coreference resolution, semantic similarity) and \textit{Text Classification} (2/20, sentiment analysis, document categorization) tasks. Then, \textit{Molecules and Proteins ML} tasks (5/20) focus on predicting molecular properties and solving biochemical modeling problems, whereas \textit{Time Series} tasks (3/20) perform forecasting over temporal data. Finally, \textit{Code} (2/20) and \textit{Math} (1/20) tasks  correspond to code generation/retrieval and mathematical reasoning, respectively.

The datasets included in \airsbench{} span both unstructured data (plain text) and structured modalities (such as tables and graphs), presenting challenges for even the most advanced models~\citep{fatemi2023talklikegraphencoding, lupidi2025source2synthsyntheticdatageneration}.

\subsection{Key Task Fields}



Table~\ref{tab:task_description} lists the core task fields of an \airsbench~task. Specifically, \texttt{name} contains the unique task name, \texttt{research\_problem} describes the problem the agent is asked to solve, \texttt{dataset} contains the HuggingFace dataset identifier,~\texttt{config} corresponds to the subset of the HuggingFace dataset used and ~\texttt{train\_split} and ~\texttt{test\_split} to the HuggingFace datasets splits used by the agent for training and evaluation, respectively. The ~\texttt{metric} field contains the metric the agent is asked to minimize or maximize (depending on the value of ~\texttt{metric\_lower\_is\_better}), while~\texttt{input\_columns} contains the dataset columns the agent can utilize to solve the research problem and  ~\texttt{scoring\_column} is the column the agent's solution will be evaluated against.
Finally, \texttt{category} describes the domain of the research problem, \texttt{sota\_paper\_title}, \texttt{sota\_paper\_url}, \texttt{sota\_year} and \texttt{sota\_venue} hold information related to the paper containing the state-of-the-art result and \texttt{sota\_score} is the state-of-the-art value of the \texttt{metric} appearing in the paper; the agent is not accessing any of the information stored in the category and sota-related fields. \\

\begin{table}
\small
\centering
\begin{NiceTabular}{lX}
\toprule
\textbf{Field} & \textbf{Value} \\
\midrule
\textcolor{blue}{\texttt{name}}              & \texttt{MathQuestionAnsweringSVAMPAccuracy} \\
\textcolor{blue}{\texttt{research\_problem}} & \texttt{Math Question Answering} \\
\textcolor{blue}{\texttt{dataset}}           & \texttt{ChilleD/SVAMP} \\
\textcolor{blue}{\texttt{metric}}            & \texttt{Accuracy} \\
\textcolor{blue}{\texttt{metric\_lower\_is\_better}} & \texttt{false} \\
\textcolor{blue}{\texttt{config}}            & \texttt{default} \\
\textcolor{blue}{\texttt{train\_split}}      & \texttt{train} \\
\textcolor{blue}{\texttt{test\_split}}       & \texttt{test} \\
\textcolor{blue}{\texttt{input\_columns}}    & \texttt{question\_concat} \\
\textcolor{blue}{\texttt{scoring\_column}}   & \texttt{Answer} \\
\textcolor{blue}{\texttt{category}}          & \texttt{Math} \\
\textcolor{blue}{\texttt{sota\_score}}       & \texttt{0.942} \\
\textcolor{blue}{\texttt{sota\_year}}        & \texttt{2026} \\
\textcolor{blue}{\texttt{sota\_venue}}       & \texttt{Frontiers of Computer Science} \\
\textcolor{blue}{\texttt{sota\_paper\_title}} & \textit{Achieving >97\% on GSM8K: Deeply Understanding the Problems Makes LLMs Better Solvers for Math Word Problems} \\
\textcolor{blue}{\texttt{sota\_paper\_url}}  & \url{https://arxiv.org/pdf/2404.14963v5} \\
\textcolor{blue}{\texttt{dataset\_paper\_url}} & \url{https://arxiv.org/abs/2103.07191} \\
\bottomrule
\end{NiceTabular}
\caption{Core fields of the \texttt{MathQuestionAnsweringSVAMPAccuracy} task stored in its \texttt{metadata.yaml} file.}
\label{tab:task_description}
\end{table}



\subsection{Task  Files}
Table~\ref{tab:task_description} describes most of the key information that forms an \airsbench~task. This information requires additional code and data to be run on the ~\airadojo~ and ~\mlgym~ scaffolds. Sections ~\ref{app:mlgym_details} and~\ref{app:airadojo_details} of the Appendix contain the system prompts of the two scaffolds. The full task specification includes a folder, whose name is the same as the task name, and a number of files that are required for the agent to solve the task within~\airadojo. The task files are organized in such a way that they can be easily and programmatically converted into files required by different harnesses; we show this by converting them into task definition files for the ~\mlgym~agentic framework.  The linked Github repository\footnote{\url{https://github.com/facebookresearch/airs-bench}} contains the \airsbench~task specifications for~\airadojo~ and~\mlgym~ along with scripts for \airadojo-to-\mlgym~format conversion and scripts for preparing the experimental environment (e.g. dataset downloads). In the remaining of this section, we describe the format and purpose of the task definition files for \airadojo.


\subsubsection{project\_description.md}
\label{sec:projectdescription}
The~\texttt{project\_description.md} file contains the instructions provided to the agent to complete the task in the form of a lengthy and appropriately structured prompt; see Appendix~\ref{app:projectdescription.md} for a full example. The prompt is divided into three sections: a description of the research problem, a description of the dataset and an explanation of the evaluation setup.

The research problem is presented in a sentence that describes the objective of the problem along with the column of the dataset that the predictions will be evaluated against. For the running example, this is:\\
``Your task is to solve math word problems. Each example presents a short story followed by a specific question. Your task is to read the text and predict the correct numerical answer. Your predictions will be scored against the \texttt{Answer} column of the test set
''


In the dataset description, we report the structure of the HuggingFace dataset. In particular, we specify  which repo the data comes from and the dataset schema (features) with an overview of the columns and their datatypes. This helps the agent understand how the data looks like even if the scaffold does not provide a lookahead function. All the data used by the agent during a task is pre-downloaded and exported to the agent's container.

Lastly, we explain to the agent how to submit the solution and how it will be scored:
the agent is expected to submit its solution in the form of a \texttt{.csv} file containing the predictions on the test split, which has the benefit of a standard output format.
We also provide the agent with the code of the evaluation script (\texttt{evaluate.py}, explained below) that contains the metric implementation and will be used to score the agent-produced \texttt{submission.csv} file against the test data.

\subsubsection{prepare.py and evaluate\_prepare.py}
\label{sec:prepare.py}

The \texttt{prepare.py} and \texttt{evaluate\_prepare.py} files contain the one-time data preparation logic for the agent to solve the problem and for the evaluation of the agent's solution, respectively. Please note that there are differences between the two settings, as the test labels need to be removed while the agent is building its solution; the two scripts take care of these requirements.

\subsubsection{evaluate.py}
\label{sec:evaluate.py}


The \texttt{evaluate.py} file is the evaluation script used to score the agent's submissions against the test data. See Appendix ~\ref{app:evaluate.py} for the evaluation script of the \texttt{MathQuestionAnsweringSVAMPAccuracy} task. The script contains three core functions: \texttt{load\_test\_set} where the script loads the test data, \texttt{evaluate} which implements the metric used to score the submissions, and \texttt{cli} which orchestrates loading of the agent's submissions and test data, running the \texttt{evaluate} method on these, and reporting the results to stdout.

\subsubsection{metadata.yaml}
\label{sec:metadata.yaml}
The \texttt{metadata.yaml} file contains all the metadata about the task (same as the fields of Table ~\ref{tab:task_description} described in detail above) along with additional requirements to run the task (like libraries used by the evaluation script that need to be installed). See Appendix ~\ref{app:config.yaml} for the \texttt{metadata.yaml} file of the \texttt{MathQuestionAnsweringSVAMPAccuracy} task.

\subsubsection{utils.py}

The \texttt{utils.py} file is an optional file to consolidate overlapping code between the \texttt{prepare.py}, \texttt{evaluate.py} and \texttt{evaluate\_prepare.py} files. Examples include normalization transforms used for data preparation or bespoke label extraction logic.

\subsubsection{Train and test datasets}

All datasets have been downloaded beforehand using the \texttt{prepare\_hf\_datasets\_text.py} script. The data required for the task is then mounted to the agent's container and prepared using the code within \texttt{prepare.py} and \texttt{evaluate\_prepare.py}. The data folder always contains a train split under a folder whose name is the value of the \texttt{train\_split} field of \texttt{metadata.yaml}  and a test split under a folder similarly named using the \texttt{test\_split} field.
For the train split, one preparation step is removing all but the relevant columns for the task (and transforming their content if needed). This is to ensure that the model does not have access to extra data that might hint to the solution. The test split contains the test labels when used by the evaluation script to score the agent's submissions, but it  does not contain the labels when accessed by the agent to  look at the structure of the test set and solve the problem.

\section{Experiments}
\label{sec:exp_design}

\subsection{Evaluation Setup}
We evaluate \airsbench~using two harnesses, \mlgym~and \airadojo. To isolate the effects of design choices associated with different harnesses, we ensured similar constraints and resources across all runs. For further details, refer to Table~\ref{app:rad-mlgym-comparison} in Appendix~\ref{sec:harness}. The greedy scaffold within \airadojo~explores several solutions through a tree-based search policy, while \mlgym~operates sequentially within one reasoning stream. Each run lasts for 24 hours with access to one H-200 GPU. We launch each run of each task at least 10 times (which we refer to as 10 ``seeds'').

Throughout \airsbench~evaluations with \mlgym~and \airadojo~harnesses, agents are allowed to access HuggingFace checkpoints, permitting the use of pretrained models. To facilitate this process and to mitigate HuggingFace rate limits, we locally cache a number of pretrained checkpoints. Note that the cache does not offer access to the latest foundational models and the most recent cached model dates back to $2021$. The full list can be found in section~\ref{app:model_cache} of the Appendix. For both \airadojo~and \mlgym~runs, agents were explicitly instructed about the existence of the cache.
Across all experiments, we do not provide agents with any information regarding the methodology or score of the SOTA paper. We hypothesize that some of the tasks in the benchmark would have benefited from more compute or time, but we kept costraints uniform across tasks to provide the agents with the same resources and push the limits of their ideation capabilities.

\subsection{Metrics and Score Aggregation}
\label{sec:metrics}

We evaluated the performance of the agents on \airsbench{} using three metrics: mean valid submission rate, average normalized score and Elo rating. The definitions of these metrics are provided below. Throughout all metrics and empirical results, we follow the terminology introduced in Section \ref{sec:agents}, for which an agent $a$ is defined as the combination of a scaffold (e.g. Greedy) and a base LLM (e.g. gpt-oss).

For each task, the agents faced the challenge of being able to \textit{submit} a valid solution, i.e. one that meets the requirements specified in the task description and yields a valid score. Our first evaluation metric is thus the mean \textbf{valid submission rate} (VSR) across tasks for an agent $a$, defined as

\begin{equation}
\overline{\text{VSR}}_a = \frac{1}{N_a} \sum_{t=1}^{N_a} \frac{valid_{a,t}}{total_{a,t}}
\label{eq:meanvalidrate}
\end{equation}

where $valid_{a,t}$ is the number of valid (successful) runs for agent $a$ on task $t$, $total_{a,t}$ is the number of total runs for agent $a$ on task $t$, $N_a$ is the number of tasks over which agent $a$ is being evaluated (i.e. the \airsbench{} tasks) and agent $a$ is a combination of a base LLM with a scaffold. The mean valid submission rate assesses the agents' capability to come up with a working solution and submit it confidently.

Producing an aggregate score for \airsbench{} is challenging due to the high diversity of tasks included: most tasks have unique metrics, and even for tasks sharing the same metric (e.g. accuracy), ranges reported in the literature for each of them may vary significantly. To aggregate heterogeneous metrics and ranges into a common scoring system, we define the \textbf{normalized score} (NS) of an agent $a$ on a task $t$ as:
\begin{equation}
    \text{NS}_{t}^a = \frac{\phi_t(s_{t}^a) - \phi_t(s^\mathrm{min}_t)}{\phi_t(s^\mathrm{sota}_t) - \phi_t(s^\mathrm{min}_t)}
    \label{eq:normscore}
\end{equation}

where $s_t^\mathrm{min}$ corresponds to the worst score observed across all seeds and all agents on task $t$, $s_t^\mathrm{sota}$ is the SOTA score on task $t$ sourced from literature, $s_{t}^a$ is the score achieved by agent $a$ on task $t$ and $\phi_t$ is a non-linear transformation. Note that $\phi_t(s_t^\mathrm{min})$ and $\phi_t(s_t^\mathrm{sota})$ will always correspond to normalized scores of $0$ and $1$, respectively.
Equation~\ref{eq:normscore} involves a two-step normalization: first, for a given task $t$, we apply a monotonic map $\phi_t$ onto the raw score $s$ achieved by the agent's submission; second, we restrict the resulting scores within the $[0,1]$ interval if the agent performs worse than SOTA and $>1$ if the agent exceeds SOTA to enable subsequent aggregation across tasks. Specifically, we employ the \textbf{march of 9s} transform\footnote{The "march of nines" is a metaphorical expression, popularized by AI researcher Andrej Karpathy, to describe the vast and non-linear engineering effort required to achieve higher levels of reliability in AI systems \citep{karpathy_dwarkesh_2025}.} as our choice of $\phi_t$, defined as
    \begin{equation}
        \phi_t(s) = -\log_{10}(|s - s^\mathrm{opt}_t|)
    \label{eq:marchof9s}
    \end{equation}
where $s^\mathrm{opt}_t$ is the overall possible optimal score for the task (e.g., $1.0$ for classification accuracy, $0.0$ for regression error), as opposed to the best score obtained or SOTA (which e.g. for accuracy would be less than 1.0). This choice of $\phi_t$ is to adjust changes of the score so that they reflect intuitive measures of progress on the benchmark: this approach treats closing e.g. the gap from $0.99$ to $0.999$ as significant as closing the gap from $0.9$ to $0.99$, since both represent a tenfold reduction in the distance to optimal.\footnote{The reader can contrast this with a simpler but less representative transform definition, such as the identity transform, for which we present results in Section~\ref{sec:additional-results} of the Appendix} When averaging normalized scores across seeds, we include both failed (i.e., the agent fails to submit a valid solution) and invalid submissions (i.e., the agent submits a solution that does not yield a numerical score) treating them as submissions with 0 normalized score.

Lastly, to quantify the relative skill level of each agent evaluated, we employ the \textbf{Elo rating system} ~\citep{elo1967proposed}. We do so by treating each agent as a player. For each \airsbench~task, we treat each pairwise comparison of the agents' scores as a game, with the agent producing a better score winning that game. If two agents do not produce a valid submission or both produce the same score, that game is considered a tie.

We estimate the agents’ ratings by fitting a Bradley–Terry (BT) model to the head-to-head outcomes, following the approach used in Chatbot Arena \citep{chiang2024chatbotarena}. The model infers latent skill parameters ($\theta_a$ for agent $a$) such that the probability of one agent $a$ outperforming another agent $b$ follows a logistic function of their skill difference:

\begin{equation}
P(a > b) = \frac{1}{1 + \exp(\theta_b - \theta_a)}
\end{equation}
We then convert the estimated skill parameters $\theta_a$ into Elo ratings using the following transformation, where N denotes the total number of evaluated agents:
\begin{equation}
R_a = \frac{400}{\ln(10)} \cdot \left[\theta_a - \frac{1}{N}\sum_{k=1}^{N} \theta_k\right] + 1000
\end{equation}
Unlike the classical Elo calculation, the BT model is order-invariant, making it well-suited for batch evaluations across multiple agents and tasks. Please note that in our setting we treat the human SOTA scores as an additional agent, the ``SOTA'' agent.

\section{Results}
\label{sec:exp_results}

We evaluate a total of 14 agents, i.e. LLM-scaffold pairs. The language models used are the Code World Model (CWM), o3-mini, gpt-oss-20b and gpt-oss-120b, GPT-4o and Devstral-Small 24B. We evaluate three scaffolds: (i) One-Shot, where the agent can attempt solving the problem only once with the same set of \airadojo~operators (by definition that would be the \textit{Draft} operator only) (ii) Greedy, where the agent performs greedy search with the \airadojo{}
operator set and (iii) ReAct, where the ReAct prompting technique implemented by~\mlgym~is powering the agent.

\begin{figure}[t]
    \centering
    \includegraphics[width=\linewidth]{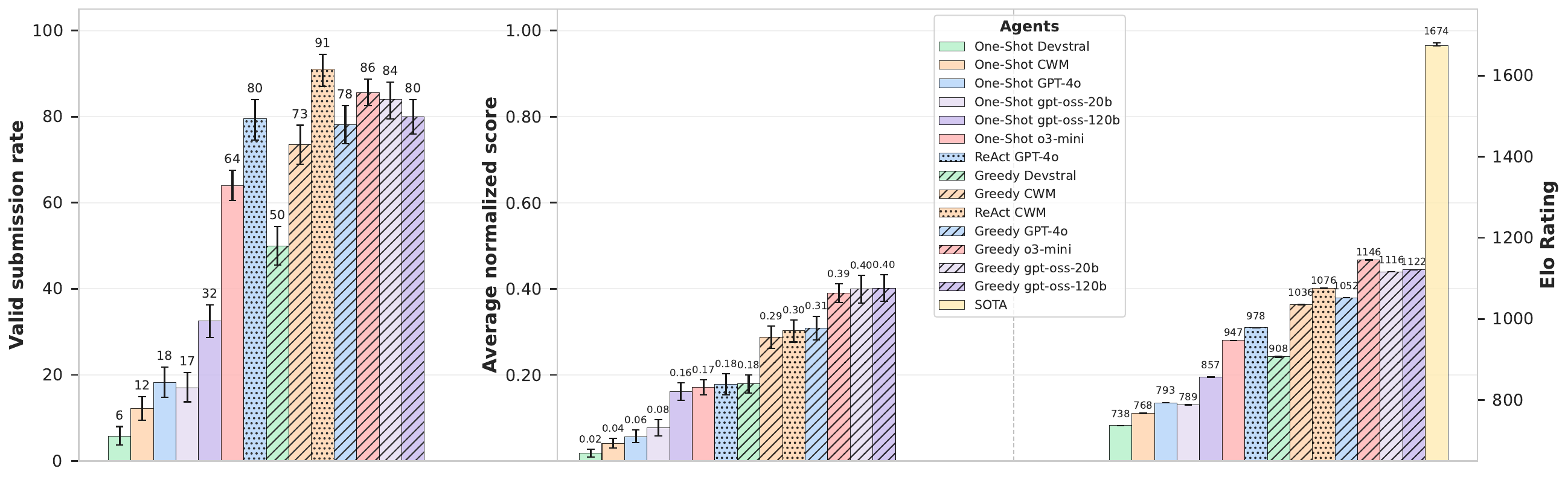}
    \caption{Overall performance of the 14 evaluated agents on the three metrics introduced in Section~5.2, namely valid submission rate, average normalized score and Elo rating. Results are ordered by increasing average normalized score.}
    \label{fig:metrics_summary}
\end{figure}

\subsection{Comparing performance across agents}
\label{sec:comparing}

Figure~\ref{fig:metrics_summary} provides an overview of the three aggregate benchmark metrics introduced above. We observe that reasoning models (e.g. gpt-oss-120b, o3-mini) perform better in both one-shot and greedy settings. However, model size seems to not affect performance with the use of test-time scaling making up for it, i.e. Greedy gpt-oss-120b is on par with Greedy gpt-oss-20b. Moreover, tree-search methods benefit agents powered by both open-source and closed-source models, as suggested by e.g. the sizeable gaps between the performances of Greedy CWM and One-Shot CWM as well as Greedy GPT-4o and One-Shot GPT-4o agents. At the same time, performance of agents backed by linear scaffolds, such as ReAct CWM and ReAct GPT4o varies; in one case the ReAct agent is better than its greedy counterpart (CWM) and in the other case worse (GPT4o). We observe that the relative ranking of agents shows similar (but not identical) trends for the average normalized score and Elo Rating performance metrics. However, the ability to submit a valid solution does not always correlate with the ability to submit a high-performing solution. This suggests that agents who are willing to take risks may achieve better performance.
Finally, the majority of agent-task combinations achieve results between the worst possible score and human SOTA, with a small but notable fraction (1.58\%) exceeding SOTA, primarily driven by greedy search strategies.

\begin{figure}[!htbp]
    \centering
    \includegraphics[width=\linewidth]{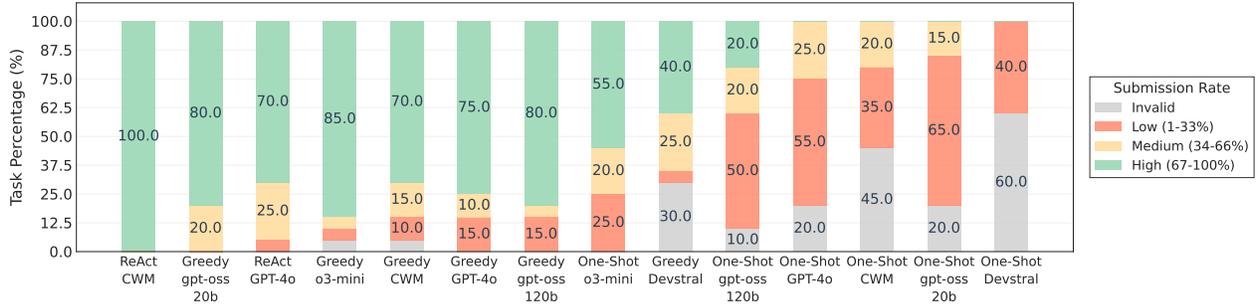}
    \vspace{-0.75cm}
    \caption{Submission rate distribution for the 14 agents tested. Each bar shows the distribution of submission rates across tasks for a given agent. The categories are defined as follows: \textit{invalid} indicates that the agent did not provide any valid submission for that task (0\% valid submissions); \textit{low (1--33\%)} indicates a valid submission for between 1\% and 33\% of seeds; \textit{medium (34--66\%)} indicates a valid submission for between 34\% and 66\% of seeds; and \textit{high (67--100\%)} indicates a valid submission for more than 66\% of seeds. Agents are sorted by the combined percentage of seeds in the \textit{medium} and \textit{high} categories, highlighting those most reliable across the benchmark.}
    \label{fig:submissionrate}
\end{figure}

We present valid submission rates for each agent in Figure~\ref{fig:submissionrate}, highlighting different ranges of valid submission rate across all tasks.
We consider four submission rate ranges: \textit{invalid} indicates that the agent failed to submit a valid solution across all its seeds;  \textit{low}/\textit{medium}/\textit{migh} correspond to the 1--33\%, 34--66\% and 67--100\%  of valid submission rates. Agents Greedy gpt-oss-120b and Greedy gpt-oss-20b lead with the smallest fractions of tasks yielding an invalid submission, at 6\% and 7\% respectively.

\begin{figure}[t]
    \centering
    \includegraphics[width=\linewidth]{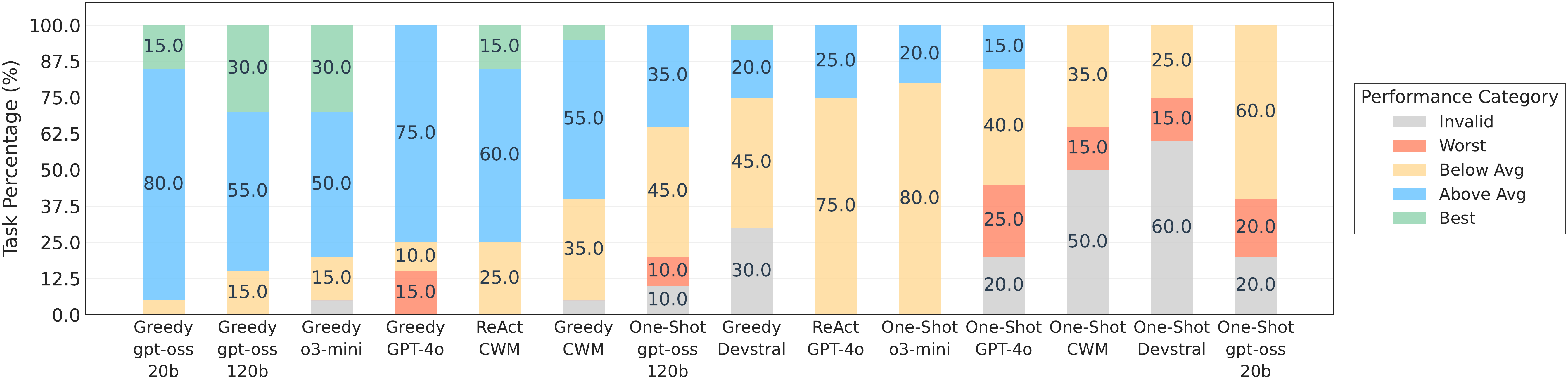}
    \vspace{-0.75cm}
    \caption{
Performance distribution for the 14 agents evaluated. Each bar represents the percentage of tasks across all seeds for which a given agent falls into one of five performance categories: \textit{invalid} (no valid submission for the task), \textit{worst} (the lowest normalized score among all agents for the task), \textit{below average} (normalized score below the mean but not the worst), \textit{above average} (normalized score above the mean but not the best), and \textit{best} (the highest normalized score for the task). Normalized scores are computed per task according to equations ~\ref{eq:normscore} and~\ref{eq:marchof9s}. Agents are sorted by the number of tasks for which they achieved the \textit{best} and \textit{above average} performances, highlighting those with the most consistent top performance across the benchmark.
}
    \label{fig:performance}
\end{figure}

A breakdown of each agent's average normalized score per task is provided in Figure~\ref{fig:performance}. For each agent, we report the percentage of tasks for which the agent yields one of five possible outcomes: (i) \textit{invalid}, the agent does not submit a solution at all; (ii) \textit{worst}, the agent produces the lowest score among all agents for that task; (iii) \textit{below average}, the agents achieves a score below the average across all agents for that task; (iv) \textit{above average}, the agent achieves a score above the average but is not the best; and (v) \textit{best}, the agent achieves the highest score among all agents for that task. This breakdown highlights the distribution of each agent's performance across the benchmark tasks. Scores are normalized according to Equation~\ref{eq:normscore}, where $\phi_t$ is the march of 9s transform.

\begin{figure}[!htbp]
    \centering
    \includegraphics[width=\linewidth]{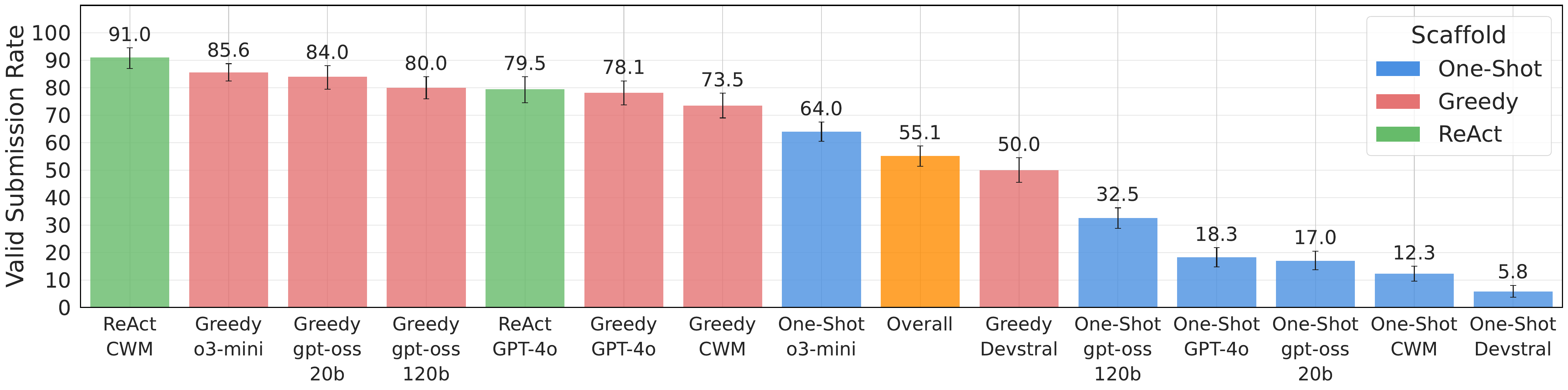}
    \vspace{-0.75cm}
    \caption{Mean valid submission rate (VSR) for the 14 agents evaluated,  with error bars indicating the $95\%$ confidence intervals. VSR is computed according to Eq.~\ref{eq:meanvalidrate}. The overall VSR across all runs and agents averages at $59.3\%$ indicating that even submitting a valid solution is non-trivial for the agents' capabilities.}
    \label{fig:meanvalidsubmissionrate}
\end{figure}

We report the mean valid submission rate defined in Equation~\ref{eq:meanvalidrate} across the \airsbench{} tasks in Figure.~\ref{fig:meanvalidsubmissionrate}. On average, only 55.1\% of the total submissions are considered valid, suggesting that even submitting a valid solution stretches the capabilities of the agents. We also observe that reasoning models and the Greedy/ReAct scaffolds offer an advantage, as performance of these agents is superior.

\begin{figure}[!htbp]
    \centering
    \includegraphics[width=0.99\linewidth]{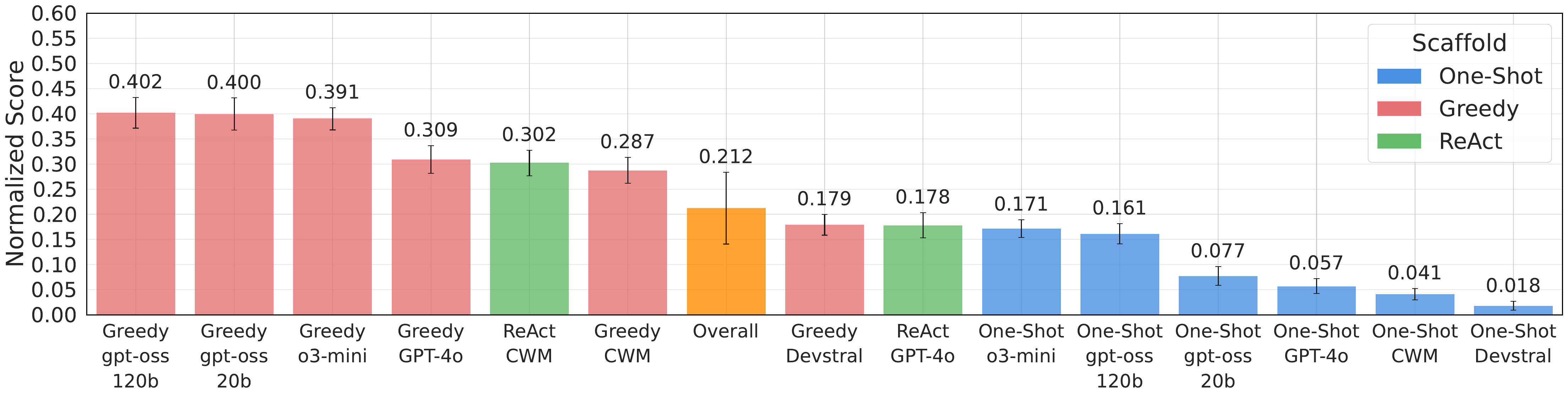}
    \vspace{-0.3cm}
    \caption{Average normalized scores for the 14 agents evaluated, with error bars indicating the $95\%$ confidence intervals. Scores are computed according to Equations~\ref{eq:normscore} and~\ref{eq:marchof9s}. The overall average normalized score across all runs and agents averages at $24.1\%$, highlighting the challenging nature of \airsbench.}
    \label{fig:meanscore}
\end{figure}

\begin{figure}[!htbp]
    \centering
    \includegraphics[width=\linewidth]{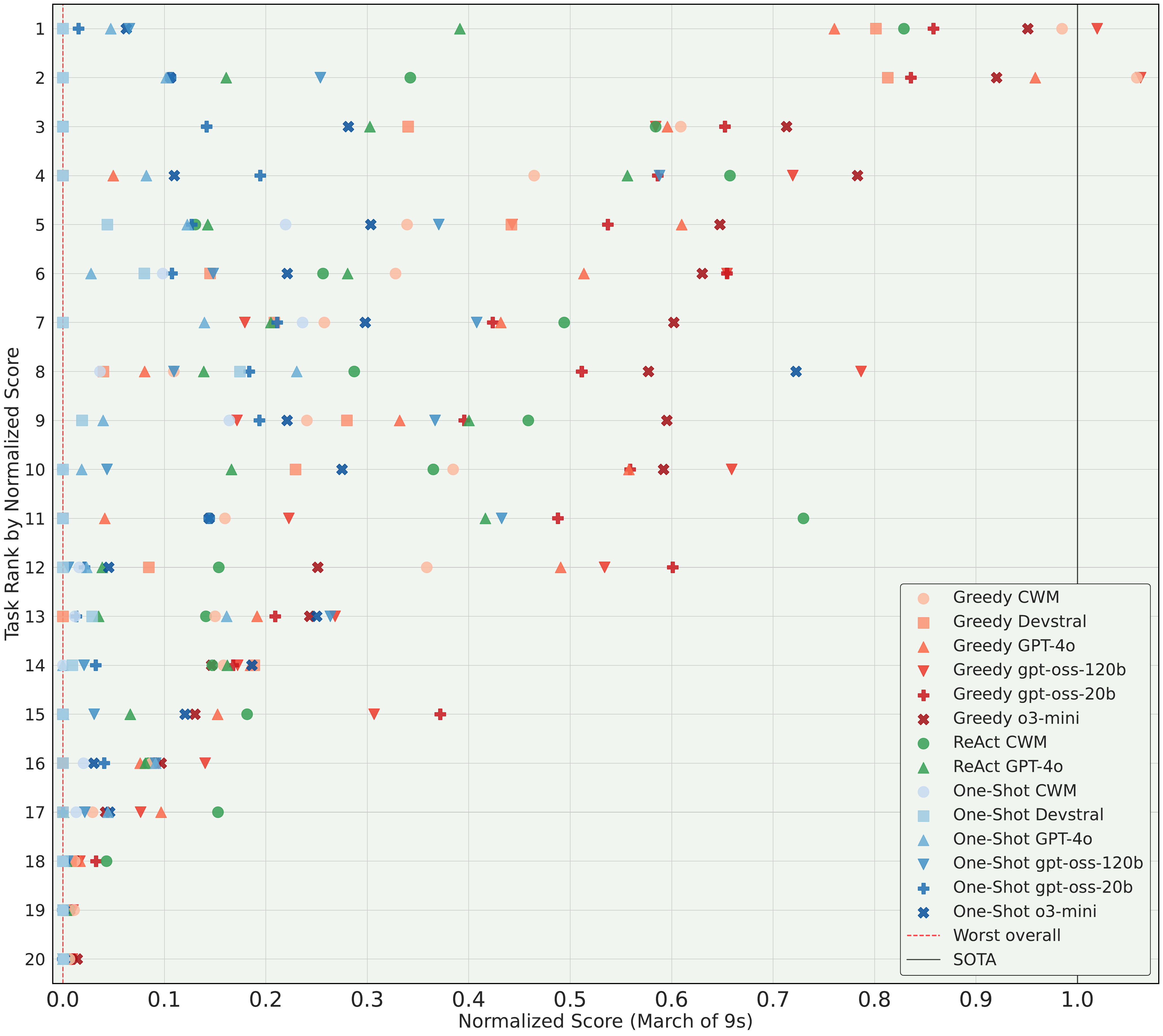}
    \caption{
    Average normalized scores with each row corresponding to an AIRS-Bench task and each point to an agent's normalized score for that task averaged across multiple seeds. For each task, the outcome of the worst-performing run is used as the baseline score. SOTA always corresponds to a normalized score of 1. Tasks are ranked in decreasing order according to the average score across all agents. See Table~\ref{tab:task-key} for the correspondence between tasks numbers on the y axis and names. 
    }
    \label{fig:normalizedscoresmarch}
\end{figure}

In Figure~\ref{fig:meanscore} we report the average normalized scores according to Equations~\ref{eq:normscore} and~\ref{eq:marchof9s}. A more detailed breakdown of the average scores per task is reported in Figure~\ref{fig:normalizedscoresmarch} with $\phi_t$ specified by Equation~\ref{eq:marchof9s}. The scores distribution in the figure reiterates the value of the~\airadojo~harness in supporting agents to develop better solutions, with Greedy scaffolds (in red) distributing closer to SOTA than One-Shot ones (in blue).

Figure~\ref{fig:normalizedscoresmarch} depicts normalized scores computed according to Equations~\ref{eq:normscore} and~\ref{eq:marchof9s}: each row corresponds to a task and each point relates to the normalized score achieved by an agent on that task and averaged across multiple seeds. For each task we average the normalized score across all agents and seeds and we use this mean normalized score to sort tasks by difficulty level. We stack tasks from the easiest to the most difficult going from top to bottom. The mapping between task numbers appearing in the y axis of Figure~\ref{fig:normalizedscoresmarch} and task names can be found in  Table~\ref{tab:task-key} of Appendix~\ref{sec:additional-results}. Tasks with points to the right of the $1.0$ line indicate that the average score of that agent on that task exceeds human SOTA, which is not necessarily the case for all tasks with at least one seed (i.e. agent run) exceeding human SOTA. Figure~\ref{fig:normalizedscoresidentity} in Appendix~\ref{sec:additional-results} similarly presents normalized scores across tasks and agents, but using in Equation~\ref{eq:normscore} the identity transform from Equation~\ref{eq:identity} (in Appendix~\ref{sec:additional-results}) instead of the march of 9s transform from Equation~\ref{eq:marchof9s}.

\begin{figure}
    \centering
    \includegraphics[width=\linewidth]{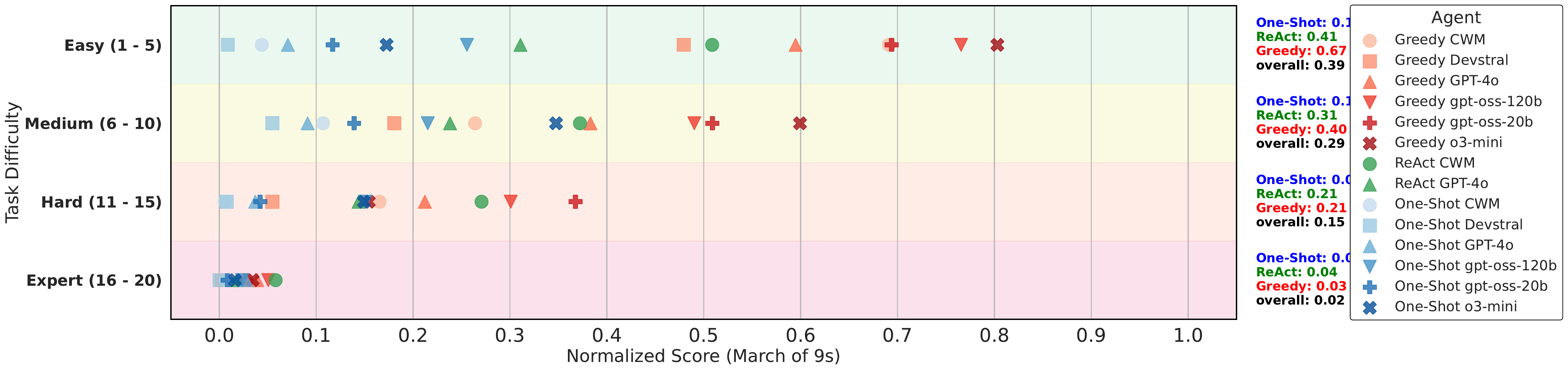}
    \caption{Normalized score per task difficulty level computed according to Equations~\ref{eq:normscore}-\ref{eq:marchof9s}. We divide the task ranking of Figure~\ref{fig:normalizedscoresmarch} into four categories with decreasing normalized scores: \textit{easy}, \textit{medium}, \textit{hard} and \textit{expert}.}
    \label{fig:categoriesmarchof9}
\end{figure}

Based on the task ranking from Figure~\ref{fig:normalizedscoresmarch}, in Figure~\ref{fig:categoriesmarchof9} we group  the \airsbench{} tasks into four groups, each containing 5 tasks, and corresponding to an increasing level of difficulty: \textit{easy} (tasks 1 to 5), \textit{medium} (tasks 6 to 10), \textit{hard} (tasks 11 to 15) and \textit{expert} (tasks 16 to 20). Here we are averaging scores across 5 tasks, i.e. each point is the average over seeds of all 5 tasks in that difficulty bucket. We also observe in Figure~\ref{fig:categoriesmarchof9} that while the normalized scores are all low and somewhat similar on the expert tasks, for the easier problems (and especially those in the easiest bucket), we see high variability between the scores that the agents achieved. The correspondence between task numbers and names is the same as the one  in Figure~\ref{fig:normalizedscoresmarch} and can be found in Table~\ref{tab:task-key} of Appendix~\ref{sec:additional-results}.

\begin{figure}[!htbp]
    \centering
    \includegraphics[width=\linewidth]{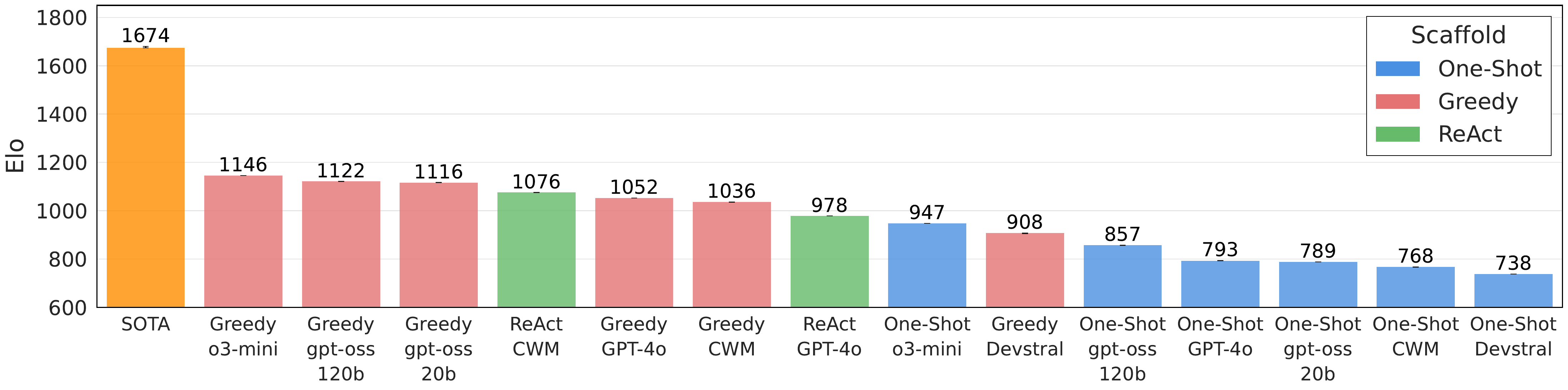}
    \vspace{-0.6cm}
    \caption{Elo ratings of all agents, estimated by fitting a Bradley--Terry model on the pairwise comparisons of agents' scores for each task. The human SOTA score is also included as an additional opponent. The Greedy scaffold outperforms other scaffolds in most cases. Bar height represents the median of a bootstrap distribution using 100 resamples, with the error bars representing the 95\% confidence intervals.}
    \label{fig:elofractionwithsota}
\end{figure}

Finally, Elo ratings including human SOTA as an additional opponent alongside our agents are reported in Figure~\ref{fig:elofractionwithsota}. The sizeable gap between the rating of the human SOTA player and the top-performing agent indicates that even the best agent is significantly below SOTA and the benchmark is very far from saturated.

\subsection{Task Inspection: Success in Beating SOTA  }

Among the runs shown in Figures \ref{fig:normalizedscoresmarch} and \ref{fig:normalizedscoresidentity}, we found cases where the agent's performance, at least in some of the seeds, was higher than the reported human SOTA, i.e. had a normalized score that was greater than $1$. Overall, we identified $4$ tasks where our agents surpass SOTA performance, as summarized across Tables \ref{tab:beat-sota-greedygptoss120b}-\ref{tab:beat-sota-greedycwm}. We examined these cases in depth to better understand the solution produced by the agent, and how it manages to outperform the human SOTA. Below, we provide a breakdown for a notable case where an agent outperforms human SOTA with an original solution.


\textbf{Greedy gpt-oss-120b on TextualClassificationSickAccuracy}. This is a Natural Language Inference (NLI) problem, and it employs the SICK (Sentences Involving Compositional Knowledge) dataset \citep{marelli2014semeval}. Given a pair of sentences, a premise and a hypothesis, the goal is to determine the relationship between them, including: \textit{entailment} (the hypothesis is true given the premise), \textit{contradiction} (the hypothesis is false given the premise); and \textit{neutral} (no conclusion can be drawn on the hypothesis given the premise). The evaluation metric is accuracy.

The SOTA solution \citep{kalouli2023curing} is achieved by fine-tuning RoBERTa \citep{liu2019roberta} on the original SICK training set and testing on the original SICK test set. The approach is straightforward: a single transformer model is fine-tuned on a specific training set, yielding a test accuracy of $90.5\%$.

The Greedy gpt-oss-120b agent, on the other hand, produces a two-level stacked ensemble that combines multiple transformer models and a meta-learner. RoBERTa-large and DeBERTa-v3-large \citep{he2023debertav3}, are independently fine-tuned on the SICK training set. Each model processes sentence pairs and outputs logits for each class. The training is performed using 5-fold stratified cross-validation, ensuring robust out-of-fold (OOF) predictions and preventing overfitting. The logits from both base models are concatenated to form a feature vector for each example. These stacked features are then used to train a logistic regression meta-learner, which learns to optimally combine the predictions of the two base models. During cross-validation, the meta-learner is trained on OOF logits, ensuring that the meta-features are unbiased and not overfitted. At test time, the base models are retrained on each fold, their test logits are averaged, and the meta-learner uses these combined logits to make the final prediction. This architecture leverages the complementary information captured by RoBERTa and DeBERTa, and the meta-learner can exploit patterns that may not be apparent to either base model alone, achieving a test accuracy of $93.1\%$.

\definecolor{headerfaintorange}{RGB}{255, 230, 200}
\begin{table}[htbp]
\centering
\footnotesize
\begin{tabularx}{\textwidth}{p{0.23\textwidth} p{0.18\textwidth} X}
\toprule
\rowcolor{headerfaintorange}
\textbf{Task} & \textbf{Score} & \textbf{Method} \\
\midrule

\makecell[l]{\texttt{TextualClassificationSick-}\\\texttt{Accuracy}} &
\begin{itemize}[nosep,leftmargin=*]
    \item \textbf{SOTA}: 0.90
    \item \textbf{Agent}: \underline{0.93}
\end{itemize} &
\begin{itemize}[nosep,leftmargin=*]
    \item \textbf{SOTA} \citep{kalouli2023curing}: Vanilla fine-tuning of RoBERTa.
    \item \textbf{Agent}: Finetuned RoBERTa-large and DeBERTa-v3-large base models using stratified cross-validation.
    Out-of-fold logits from both models employed to train a logistic-regression meta-learner that combines base models' logits. Finally, base models retrained on all folds and meta-learner was used to produce final predictions.
\end{itemize} \\
\midrule

\makecell[l]{\texttt{TextualSimilaritySick-}\\\texttt{SpearmanCorrelation}} &
\begin{itemize}[nosep,leftmargin=*]
    \item \textbf{SOTA}: 0.85
    \item \textbf{Agent}: \underline{0.89}
\end{itemize} &
\begin{itemize}[nosep,leftmargin=*]
    \item \textbf{SOTA} \citep{huang2024cosent}: Finetuned RoBERTa-large and novel loss function (CoSENT).
    \item \textbf{Agent}: RoBERTa-base and RoBERTa-large finetuned to predict similarity scores. Produces similarity scores using cosine similarity of frozen Sentence-BERT. Used cross-validation to learn weights for averaging the similarity scores produced by all three models.
\end{itemize} \\

\bottomrule
\end{tabularx}

\caption{\airsbench{} tasks where the \textbf{Greedy gpt-oss-120b} agent surpassed human SOTA performances in at least one run, and achieves the best overall score. The left column displays the name of the task $t$; the middle column shows SOTA and the raw agent scores $s^\mathrm{sota}_t$ and $s_t^a$, respectively; the right column briefly summarises and compares the SOTA and the Agent solutions.}
\label{tab:beat-sota-greedygptoss120b}

\end{table}

\definecolor{faintblue}{RGB}{237, 243, 252}

\begin{table}[htbp]
\centering
\footnotesize
\begin{tabularx}{\textwidth}{p{0.23\textwidth} p{0.18\textwidth} X}
\toprule
\rowcolor{faintblue}
\textbf{Task} & \textbf{Score} & \textbf{Method} \\
\midrule
\makecell[l]{\texttt{CoreferenceResolution-}\\\texttt{WinograndeAccuracy}} &
\begin{itemize}[nosep,leftmargin=*]
    \item \textbf{SOTA}: 0.85
    \item \textbf{Agent}: \underline{0.88}
\end{itemize} &
\begin{itemize}[nosep,leftmargin=*]
    \item \textbf{SOTA} \citep{lin2020tttttackling}: Fine-tune T5-3B in a text-to-text setup, scoring answer options by output token probabilities (“entailment” vs. “contradiction”).
    \item \textbf{Agent}: Vanilla finetuning of DeBERTa-v3-large with classifier head.
\end{itemize} \\
\bottomrule
\end{tabularx}
\caption{\airsbench{} tasks where the \textbf{Greedy gpt-oss-20b} agent surpassed human SOTA performances in at least one run, and achieves the best overall score. The left column displays the name of the task $t$; the middle column shows SOTA and the raw agent scores $s^\mathrm{sota}_t$ and $s_t^a$, respectively; the right column briefly summarises and compares the SOTA and the Agent solutions.}
\label{tab:beat-sota-greedygptoss20b}

\end{table}

\definecolor{faintgreen}{RGB}{230, 255, 230}
\begin{table}[htbp]
\centering
\footnotesize
\begin{tabularx}{\textwidth}{p{0.23\textwidth} p{0.18\textwidth} X}
\toprule
\rowcolor{faintgreen}
\textbf{Task} & \textbf{Score} & \textbf{Method} \\
\midrule
\makecell[l]{\texttt{TimeSeriesForecasting-}\\\texttt{RideshareMAE}} &
\begin{itemize}[nosep,leftmargin=*]
    \item \textbf{SOTA}: 1.185
    \item \textbf{Agent}: \underline{1.153}
\end{itemize} &
\begin{itemize}[nosep,leftmargin=*]
    \item \textbf{SOTA} \citep{gong2025bridging}: General transformer-based time-series foundation model (not finetuned on this dataset).
    \item \textbf{Agent}: Trains a Bi-directional GRU
\end{itemize} \\
\bottomrule
\end{tabularx}
\caption{\airsbench{} tasks where the \textbf{Greedy CWM} agent surpassed human SOTA performances in at least one run, and achieves the best overall score. The left column displays the name of the task $t$; the middle column shows SOTA and the raw agent scores $s^\mathrm{sota}_t$ and $s_t^a$, respectively; the right column briefly summarises and compares the SOTA and the Agent solutions.}
\label{tab:beat-sota-greedycwm}

\end{table}

\section{Conclusion}
\label{sec:discussion}

We introduced \airsbench, a benchmark designed to rigorously evaluate the autonomous research capabilities of LLM agents in machine learning. Our benchmark covers 20 diverse, non-contaminated tasks spanning multiple domains, and is specifically constructed to assess agents across the full research workflow---from ideation and methodology design to experimentation and iterative refinement---without access to baseline code.
Our results indicate a high variability in task performance, depending on both the LLM that the agent uses, and the harness it is based on. For most tasks, even the best performing agent is still significantly behind the human SOTA, showing that the benchmark is far from saturated. For a few tasks, our top agents managed to identify a solution outperforming the human SOTA. It would be interesting to see how much AI research agents can push the state-of-the-art further.\footnote{Note that even in the cases where the agent outperforms human SOTA, it is still interesting to see how far ahead the agent's solution is over the human baseline. Hence, tasks where the agent normalized score exceeds 1.0 are still interesting benchmark tasks.}

We gathered a number of useful takeaways during building of~\airsbench{} and evaluating a number of agents across its tasks:

\begin{itemize}
    \item Gaps in community infrastructure: within the current state of AI research, the task of tracking up-to-date SOTA became more challenging than ever. Both the growing amounts of paper submissions,\footnote{\url{https://forum.cspaper.org/topic/76/submission-tsunami-at-neurips-2025-is-peer-review-about-to-collapse/2}} the high compute cost of reproducing experiments on large models and the lack of unified platforms to represent results    contribute to the situation. A new shared space with standardized format, updates, and machine-readable configurations for all published machine learning research is needed.
    \item Performance gaps: several main factors cause the performance of the agents to be lower than it could be. The combinations of base LLMs with different scaffolds can lead to 1) problems with formatting, specifically submitting the correct solution after the experiments, 2) problems with saving intermediate results, 3) performance deterioration on main capabilities due to the context overflow. Longer agentic traces also lead to increased probabilities of misaligned behaviours and accumulated issues around code edits and debugging, which make it difficult to adhere to the methodology of the ML experimentation.
    \item Human bottlenecks: for the work to be scaled across new domains and a bigger volume of tasks, automatic task onboarding pipelines will be needed. Current human validation procedures prevent the expansion at scale.
    \item Role of restrictions: The ML benchmarks for AI Agents tend to be computationally costly, both on the training and inference side. Given the significant resource constraints faced in agentic evaluations---such as computational costs, time limits, and token usage---we acknowledge the possible role of restrictions in the obtained results. Benchmark methodology commonly faces the choice of either evaluating the systems in the very well-defined restricted conditions or lifting most of them and comparing the best obtained results. Although we adhere to the first choice for the sake of future extensive ablations, lifting certain restrictions could enable more flexible and efficient agent behaviors in the future.

\end{itemize}

Our results demonstrate that scaffold design significantly impacts agent solution quality, highlighting opportunities to improve performance through algorithms that better leverage test-time compute. We release AIRS-Bench to help identify performance gaps in AI research agents and catalyze the development of better methods for accelerating scientific progress. As agentic capabilities advance, continued benchmark development will be essential. We hope this benchmark fosters transparency, reproducibility, and rigorous, standardized evaluation of LLM agents in frontier research contexts.

\bibliographystyle{plainnat} 
\bibliography{paper}

\newpage
\appendix
\section*{Appendix}

\section{Task Selection}
\label{sec:selection-process}

We constructed \airsbench{} by downsampling a pool $\mathcal{F}$ of approximately 100 tasks to a representative subset $\mathcal{S}$ of 20 tasks. The reduction to 20 tasks was implemented to substantially decrease GPU requirements and enable faster benchmarking. The \airsbench{} subset was selected to closely mirror the full pool $\mathcal{F}$ according to three key criteria:
\begin{itemize}
    \item \textbf{Agent performance}: each agent's average score on \airsbench{} is as close as possible to their average score on the full benchmark.
    \item \textbf{Category distribution}: the proportion of tasks from each of the 7 categories in \airsbench{} is as close as possible to that of the full pool.
    \item \textbf{Relative ranking fidelity}: the ranking of agents by performance is preserved between \airsbench{} and the full pool.
\end{itemize}

For each agent $a$, we compute the mean normalized score $\overline{\text{NS}}_\mathcal{F}^a$ over $\mathcal{F}$ and $\overline{\text{NS}}^a_{\mathcal{S}}$ over \airsbench{}, where

\begin{equation}
    \overline{\text{NS}}^a_{\mathcal{F}} = \frac{1}{|\mathcal{F}|}\sum_{t \in \mathcal{F}} \text{NS}_t^a \qquad
    \overline{\text{NS}}^a = \frac{1}{|\mathcal{S}|}\sum_{t \in \mathcal{S}} \text{NS}_t^a
\end{equation} where the normalized score $\text{NS}_t^a$ is defined in Equation~\ref{eq:normscore} (see Section~\ref{sec:metrics}) and is the performance score of agent $a$ on task $t$. The \airsbench{} subset is selected to minimize the mean absolute error (MAE) between $\overline{\text{NS}}^a_\mathcal{F}$ and $\overline{\text{NS}}^a_{\mathcal{S}}$ across all agents:

\begin{equation}
    \mathrm{MAE} = \frac{1}{|A|} \sum_{a \in A} \left| \overline{\text{NS}}^a_\mathcal{F} - \overline{\text{NS}}^a_{\mathcal{S}} \right|
\end{equation}

where $A$ is the set of all agents. To ensure representative coverage, $\mathcal{F}$ is partitioned into four difficulty bands ($easy$, $medium$, $hard$, $expert$), based on their relative ranking by average normalised score (see Section \ref{sec:comparing}) and each containing approximately 25 tasks. \airsbench{} is constructed by sampling a fixed number of tasks from each band. Four candidate difficulty band distributions were evaluated:
\begin{itemize}
    \item \textbf{Uniform}: $5$ tasks each from \textit{easy}, \textit{medium}, \textit{hard}, and \textit{expert} bands ($\{5, 5, 5, 5\}$).
    \item \textbf{Medium-skewed}: $4$ \textit{easy}, $7$ \textit{medium}, $5$ \textit{hard}, and $4$ \textit{expert} tasks ($\{4, 7, 5, 4\}$).
    \item \textbf{Center-skewed}: $4$ \textit{easy}, $6$ \textit{medium}, $6$ \textit{hard}, and $4$ \textit{expert} tasks ($\{4, 6, 6, 4\}$).
    \item \textbf{Medium-heavy}: $3$ \textit{easy}, $8$ \textit{medium}, $6$ \textit{hard}, and $3$ \textit{expert} tasks ($\{3, 8, 6, 3\}$).
\end{itemize}
The final band allocation was selected by choosing the configuration that minimized the mean absolute error (MAE) between agent scores on the subset and the full benchmark. The search for the optimal subset was performed using three subset selection algorithms:
\begin{itemize}
    \item \textbf{Random search}: samples ten thousand candidate subsets, each respecting the band constraints, and retains the subset with the lowest MAE.
    \item \textbf{Simulated annealing}: iteratively swaps tasks within bands, accepting both improvements and, with decreasing probability, worse solutions to escape local minima.
    \item \textbf{Genetic algorithm}: evolves a population of candidate subsets through tournament selection, single-point crossover ($p=0.7$), and mutation ($p=0.2$) minimizing MAE over generations.
\end{itemize}

Across all 12 possible (\{algorithm\} $\times$\{band distribution\}) combinations, the best-performing configuration was obtained using the genetic algorithm with a medium-skewed band allocation, achieving a minimum MAE of $4.0\times10^{-3}$. Other competitive configurations included both uniform and skewed band distributions, with MAE values ranging from $4.6\times10^{-3}$ to $7.9\times10^{-3}$.

To validate the fidelity of \airsbench{}, we compared agent mean normalized scores and their 95\% confidence intervals on \airsbench{} versus the full pool $\mathcal{F}$.  The results demonstrate that agent rankings and score gaps are faithfully preserved, with overall mean scores and confidence intervals nearly identical between the two sets: the difference in average score between the subset and the full benchmark never exceeds 0.02 in absolute value. This confirms that the selection criterion and stratified sampling approach yield a lightweight benchmark that maintains the discriminative power and ranking structure of the original pool, while allowing efficient evaluation.

\section{Additional Results}
\label{sec:additional-results}

We consider an additional normalized score (see Eq.~\ref{eq:normscore}) which employs the identity transform

    \begin{equation}
        \phi_t(s) = \mathcal{I}(s) = s
        \label{eq:identity}
    \end{equation}

\begin{figure}[!htbp]
    \centering
    \includegraphics[width=\linewidth]{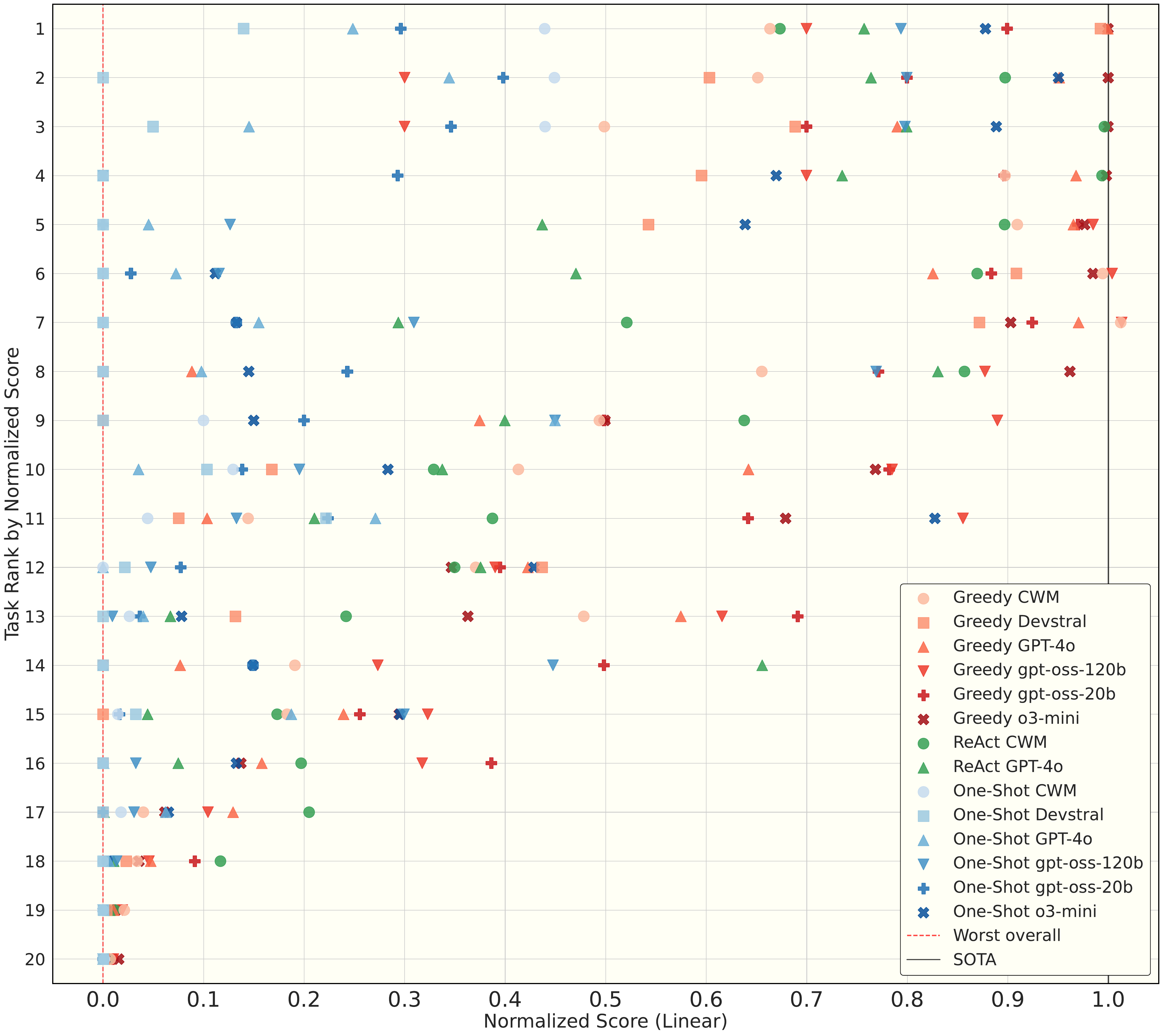}
    \caption{
    Normalized score per task averaged over seeds, computed according to Equations~\ref{eq:normscore}-~\ref{eq:identity}. For each task, the outcome of the worst-performing run is used as the baseline score $s^{\textrm{min}}_t$. SOTA always corresponds to a normalized score of 1. Tasks are ranked in decreasing order according to the average score across all agents. See Table~\ref{tab:task-key} for the correspondence between tasks ranking and name.
    }
    \label{fig:normalizedscoresidentity}
\end{figure}

\begin{figure}
    \centering
    \includegraphics[width=\linewidth]{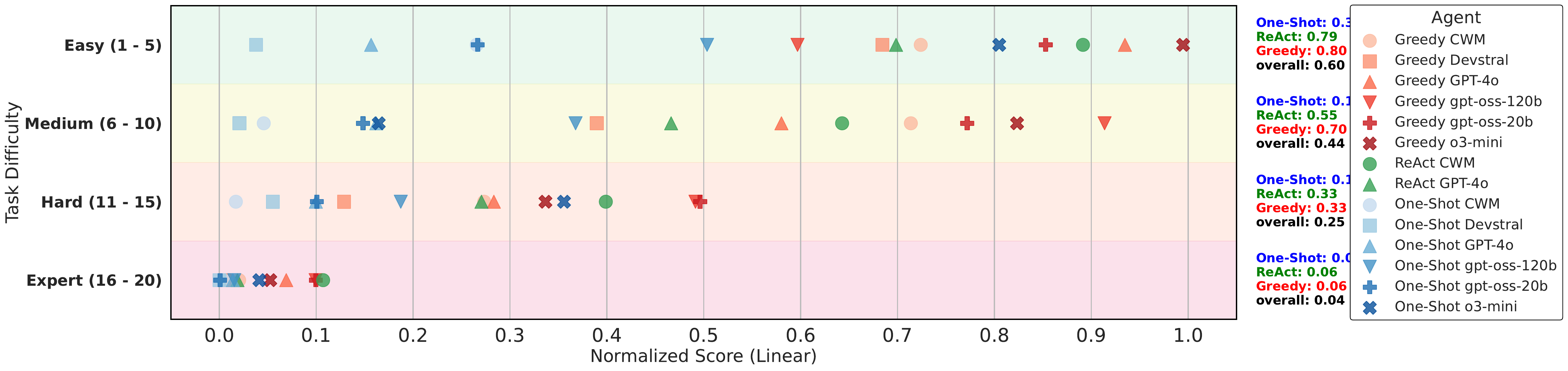}
    \caption{Normalized score per task difficulty level computed according to Equations~\ref{eq:normscore}-\ref{eq:identity}. We divide the task ranking of Figure~\ref{fig:normalizedscoresidentity} into four categories with decreasing normalized scores: \textit{easy}, \textit{medium}, \textit{hard} and \textit{expert}.}
    \label{fig:categoriesidentity}
\end{figure}

In contrast to Eq.~\ref{eq:marchof9s}, this approach directly uses raw scores. With this transform, the normalized score $\text{NS}_t^a$ linearly reflects the agent's progress between the worst observed solution and the human SOTA for each task. This is simple and interpretable, but may not always reflect meaningful progress when the metric is highly non-linear or when the gap to the optimal score is very small. For completeness, we show the average normalized score using this transform in Figure~\ref{fig:normalizedscoresidentity}, and provide a breakdown of the scores by difficulty level in Figure~\ref{fig:categoriesidentity}.

For better readability, Table~\ref{tab:task-key} provides the mapping between task numbers shown on the y-axis of Figures~\ref{fig:normalizedscoresmarch}-\ref{fig:normalizedscoresidentity} and their corresponding task names, as well as their average score across all seeds and agents.

\newpage

\definecolor{easy}{HTML}{cbeed7} 
\definecolor{medium}{HTML}{f3f3b7} 
\definecolor{hard}{HTML}{ffd1c2} 
\definecolor{expert}{HTML}{f7b6cd} 

\setlength{\tabcolsep}{1.5pt}
\renewcommand{\arraystretch}{1.4}
\captionsetup[longtable]{justification=centering}
\rowcolors{2}{white}{white}

\begin{longtable}{%
  >{\footnotesize}r
  >{\arraybackslash\footnotesize}p{0.41\textwidth}
  >{\footnotesize}r
  >{\arraybackslash\footnotesize}p{0.41\textwidth}
  >{\footnotesize}r
}

\toprule
\label{tab:task-key}
\centering \textbf{Rank} & \footnotesize\textbf{Task (March of 9's)} & \(\mathbf{NS_t^a}\) & \footnotesize\textbf{Task (Identity)} & \(\mathbf{NS_t^a}\) \\
\midrule
\rowcolor{easy!15} 1 & TextualClassificationSickAccuracy                  & 0.48 & R2AbsMolecularPropertyPredictionQm9MAE & 0.68 \\
\rowcolor{easy!15} 2 & TextualSimilaritySickSpearmanCorrelation           & 0.48 & U0MolecularPropertyPredictionQm9MAE & 0.64 \\
\rowcolor{easy!15} 3 & CvMolecularPropertyPredictionQm9MAE  & 0.34 & GMolecularPropertyPredictionQm9MAE & 0.60 \\
\rowcolor{easy!15} 4 & TimeSeriesForecastingSolarWeeklyMAE                & 0.34 & CvMolecularPropertyPredictionQm9MAE & 0.55 \\
\rowcolor{easy!15} 5 & R2AbsMolecularPropertyPredictionQm9MAE & 0.32 & GraphRegressionZincMae                        & 0.54 \\

\hline

\rowcolor{medium!15} 6 & SentimentAnalysisYelpReviewFullAccuracy            & 0.30 & TextualClassificationSickAccuracy             & 0.52 \\
\rowcolor{medium!15} 7 & U0MolecularPropertyPredictionQm9MAE  & 0.29 & TextualSimilaritySickSpearmanCorrelation      & 0.52 \\
\rowcolor{medium!15} 8 & ReadingComprehensionSquadExactMatch                & 0.28 & TimeSeriesForecastingSolarWeeklyMAE           & 0.45 \\
\rowcolor{medium!15} 9 & GMolecularPropertyPredictionQm9MAE   & 0.28 & TimeSeriesForecastingKaggleWebTrafficMASE     & 0.37 \\
\rowcolor{medium!15} 10 & GraphRegressionZincMae                             & 0.28 & SentimentAnalysisYelpReviewFullAccuracy       & 0.36 \\

\hline

\rowcolor{hard!15} 11 & TimeSeriesForecastingRideshareMAE                  & 0.20 & ReadingComprehensionSquadExactMatch           & 0.34 \\
\rowcolor{hard!15} 12 & CoreferenceResolutionWinograndeAccuracy            & 0.19 & CoreferenceResolutionSuperGLUEWSCAccuracy     & 0.26 \\
\rowcolor{hard!15} 13 & QuestionAnsweringDuoRCAccuracy                     & 0.14 & CoreferenceResolutionWinograndeAccuracy       & 0.24 \\
\rowcolor{hard!15} 14 & CoreferenceResolutionSuperGLUEWSCAccuracy          & 0.11 & TimeSeriesForecastingRideshareMAE             & 0.23 \\
\rowcolor{hard!15} 15 & QuestionAnsweringEli5Rouge1                        & 0.10 & QuestionAnsweringDuoRCAccuracy                & 0.17 \\

\hline

\rowcolor{expert!15} 16 & TimeSeriesForecastingKaggleWebTrafficMASE          & 0.07 & QuestionAnsweringEli5Rouge1                   & 0.10 \\
\rowcolor{expert!15} 17 & CodeRetrievalCodeXGlueMRR                          & 0.04 & CodeRetrievalCodeXGlueMRR                     & 0.05 \\
\rowcolor{expert!15} 18 & MathQuestionAnsweringSVAMPAccuracy                 & 0.01 & MathQuestionAnsweringSVAMPAccuracy            & 0.03 \\
\rowcolor{expert!15} 19 & QuestionAnsweringFinqaAccuracy                     & 0.00 & QuestionAnsweringFinqaAccuracy                & 0.01 \\
\rowcolor{expert!15} 20 & CodeGenerationAPPSPassAt5                          & 0.00 & CodeGenerationAPPSPassAt5                     & 0.00 \\

\bottomrule

\caption{ \centering Side-by-side ranking of tasks by average normalized score with \textbf{march of 9's} and \textbf{identity} transforms, as reported in Figure~\ref{fig:normalizedscoresmarch}, and Figure~\ref{fig:normalizedscoresidentity}, respectively. Color represents progressively harder tasks, from \textit{easy} to \textit{expert}.}
\end{longtable}

\subsection{Task run files}
\subsubsection{project\_description.md}
\label{app:projectdescription.md}

\begin{tcolorbox}[
  colback=md-bg,
  colframe=md-bg,
  boxrule=0pt,
  left=10pt,
  right=10pt,
  top=10pt,
  bottom=10pt,
  sharp corners,
  enhanced
]
{\small\color{md-text}
{\large\bfseries\color{md-heading}\texttt{Overview}}

\vspace{0.3em}

{\normalsize\bfseries\color{md-heading}\texttt{Task Description}}

\vspace{0.2em}

This is a Machine Learning project and your goal is to build a model that solves the project's TASK following the instructions provided below.

\vspace{0.3em}

\hspace{1em}\texttt{\color{md-code}{TASK: Your task is to solve math word problems. Each example presents a short story followed by a specific question. Your task is to read the text and predict the correct numerical answer. Your predictions will be scored against the \texttt{Answer} column of the test set.}}

\vspace{0.6em}

{\large\bfseries\color{md-heading}\texttt{Data}}

\vspace{0.3em}

{\normalsize\bfseries\color{md-heading}\texttt{Dataset Structure}}

\vspace{0.2em}

The default config of \texttt{SVAMP} has the following structure. Here is a description of the contents of each column including their name, what they contain, and the data type:

\begin{verbatim}
{
  "ID": string,               # Unique identifier for the problem
  "Body": string,             # The narrative or context text describing the scenario
  "Question": string,         # The actual math question asked based on the Body
  "Equation": string,         # The ground-truth equation used to compute the answer
  "Answer": string,           # The correct numerical solution to the problem (as text)
  "Type": string,             # The problem category/type. Is one of ["Subtraction", "Addition",
                              # "Common-Division", "Multiplication"].
  "question_concat": string   # Concatenation of Body and Question into one text field
}
\end{verbatim}

An example entry in the train dataset contains the following:
\begin{itemize}
  \item \texttt{ID}: "chal-777"
  \item \texttt{Body}: "There are 87 oranges and 290 bananas in Philip's collection. If the bananas are organized into 2 groups and oranges are organized into 93 groups"
  \item \texttt{Question}: "How big is each group of bananas?"
  \item \texttt{Equation}: "( 290.0 / 2.0 )"
  \item \texttt{Answer}: "145"
  \item \texttt{Type}: "Common-Division"
  \item \texttt{question\_concat}: "There are 87 oranges and 290 bananas in Philip's collection. If the bananas are organized into 2 groups and oranges are organized into 93 groups How big is each group of bananas?"
\end{itemize}

\textbf{Note:} The scoring columns (\texttt{Equation}, \texttt{Answer}, \texttt{Type}) have been removed from the test data. Ensure you preserve the original row order of the test set when producing predictions.

Train data can be loaded using:
\begin{verbatim}
from datasets import load_from_disk
train_dataset = load_from_disk('./data/train')
\end{verbatim}

Test data can be loaded using:
\begin{verbatim}
from datasets import load_from_disk
test_dataset = load_from_disk('./data/test')
\end{verbatim}

\vspace{0.6em}

{\large\bfseries\color{md-heading}\texttt{Submission File}}
\vspace{0.2em}

The submission file should contain the selected answers for the test set. The submission file should be a \texttt{.csv} file named \colorbox{md-code-bg}{\texttt{\color{md-code}submission.csv}} with the following header:
\begin{verbatim}
Answer
\end{verbatim}
And it should be of shape \colorbox{md-code-bg}{\texttt{\color{md-code}(300,1)}} — one prediction per row, aligned with the test set order. Each row must contain the numerical answer to the question. Do not include any extra columns or indices.

\vspace{0.6em}

{\large\bfseries\color{md-heading}\texttt{Evaluation}}
\vspace{0.2em}

The evaluation will be performed on the \colorbox{md-code-bg}{\texttt{\color{md-code}submission.csv}} file you have submitted using the Accuracy metric.

Evaluation will be done using the following script:
}
\end{tcolorbox}

\vspace{0.3em}
\begin{lstlisting}[
    language=Python,
    basicstyle=\small\ttfamily\color{monokai-fg},
    backgroundcolor=\color{monokai-bg},
    frame=single,
    rulecolor=\color{monokai-bg},
    breaklines=true,
    breakatwhitespace=true,
    numbers=left,
    numberstyle=\tiny\color{monokai-comment},
    stepnumber=1,
    xleftmargin=1.5em,
    framexleftmargin=1em,
    columns=flexible,
    keepspaces=true,
    showstringspaces=false,
    tabsize=4,
    texcl=false,
    mathescape=false,
    escapechar=\@,
    upquote=true,
    aboveskip=0.5em,
    belowskip=0.5em,
    lineskip=-0.5pt,
    keywordstyle=\color{monokai-keyword}\bfseries,
    stringstyle=\color{monokai-string},
    commentstyle=\color{monokai-comment}\itshape,
    identifierstyle=\color{monokai-fg},
    emphstyle=\color{monokai-function},
    emph={load_test_set, evaluate, strip_cot, _normalize, _cli},
    emphstyle={[2]\color{monokai-class}},
    emph={[2]argparse, json, numpy, np, pandas, pd, torch, load_dataset, load_from_disk, ArgumentParser, ValueError}
]
#!/usr/bin/env python3
import argparse, json
import numpy as np
import pandas as pd
from datasets import load_from_disk


def load_test_answers(path='./data/test_with_labels'):
    """
    Load answers for evaluation (string labels).
    """
    ds = load_from_disk(path)
    if "Answer" not in ds.features:
        raise RuntimeError(
            "Test data does not contain 'Answer'. "
            "Use a labeled copy like './data/test_with_labels'."
        )
    answers = list(ds["Answer"])
    return answers


def evaluate(predictions, labels):
    """
    Returns a dict of metric_name -> value
    """
    correct = np.fromiter(
        (int(p) == int(y) for p, y in zip(predictions, labels)), dtype=bool
    )
    return {"Accuracy": float(correct.mean())}


def _cli():
    p = argparse.ArgumentParser(
        description="Evaluate predictions for SVAMP test split using Accuracy."
    )
    p.add_argument("--submission-file", default="submission.csv",
                   help="Path to CSV file containing predictions.")
    a = p.parse_args()

    print("Loading test set labels...")
    labels = load_test_answers()
    n_test_samples = len(labels)
    print(f"Loaded {n_test_samples} labels.")

    print(f"Loading predictions from: {a.submission_file}")
    try:
        # Assuming CSV has no header and contains only prediction values
        # Adjust if your submission format is different (e.g., has headers, specific columns)
        submission_df = pd.read_csv(a.submission_file, header=0)
        preds = submission_df.values.squeeze()
        if preds.shape[0] != n_test_samples:
            raise ValueError(
                f"Submission file row count ({preds.shape[0]}) "
                f"does not match test set size ({n_test_samples})."
            )
    except FileNotFoundError:
        p.error(f"Submission file not found: {a.submission_file}")
    except Exception as e:
        p.error(f"Error loading submission_file: {e}")

    print("Evaluating predictions...")
    result = evaluate(preds, labels)

    print("\n--- EVALUATION RESULT ---")
    print(json.dumps(result, indent=2))


if __name__ == '__main__':
    _cli()
\end{lstlisting}

\subsubsection{evaluate.py}
\label{app:evaluate.py}

\begin{lstlisting}[
    language=Python,
    basicstyle=\small\ttfamily\color{monokai-fg},
    backgroundcolor=\color{monokai-bg},
    frame=single,
    rulecolor=\color{monokai-bg},
    breaklines=true,
    breakatwhitespace=true,
    numbers=left,
    numberstyle=\tiny\color{monokai-comment},
    stepnumber=1,
    xleftmargin=1.5em,
    framexleftmargin=1em,
    columns=flexible,
    keepspaces=true,
    showstringspaces=false,
    tabsize=4,
    texcl=false,
    mathescape=false,
    escapechar=\@,
    upquote=true,
    aboveskip=0.5em,
    belowskip=0.5em,
    lineskip=-0.5pt,
    keywordstyle=\color{monokai-keyword}\bfseries,
    stringstyle=\color{monokai-string},
    commentstyle=\color{monokai-comment}\itshape,
    identifierstyle=\color{monokai-fg},
    emphstyle=\color{monokai-function},
    emph={load_test_set, evaluate, strip_cot, _normalize, _cli},
    emphstyle={[2]\color{monokai-class}},
    emph={[2]argparse, json, numpy, np, pandas, pd, torch, load_dataset, load_from_disk, ArgumentParser, ValueError}
]
#!/usr/bin/env python3
import argparse, json
import numpy as np
import pandas as pd
from datasets import load_from_disk


def load_test_answers(path='./data/test_with_labels'):
    """
    Load answers for evaluation (string labels).
    """
    ds = load_from_disk(path)
    if "Answer" not in ds.features:
        raise RuntimeError(
            "Test data does not contain 'Answer'. "
            "Use a labeled copy like './data/test_with_labels'."
        )
    answers = list(ds["Answer"])
    return answers


def evaluate(predictions, labels):
    """
    Returns a dict of metric_name -> value
    """
    correct = np.fromiter(
        (int(p) == int(y) for p, y in zip(predictions, labels)), dtype=bool
    )
    return {"Accuracy": float(correct.mean())}


def _cli():
    p = argparse.ArgumentParser(
        description="Evaluate predictions for SVAMP test split using Accuracy."
    )
    p.add_argument("--submission-file", default="submission.csv",
                   help="Path to CSV file containing predictions.")
    a = p.parse_args()

    print("Loading test set labels...")
    labels = load_test_answers()
    n_test_samples = len(labels)
    print(f"Loaded {n_test_samples} labels.")

    print(f"Loading predictions from: {a.submission_file}")
    try:
        # Assuming CSV has no header and contains only prediction values
        # Adjust if your submission format is different (e.g., has headers, specific columns)
        submission_df = pd.read_csv(a.submission_file, header=0)
        preds = submission_df.values.squeeze()
        if preds.shape[0] != n_test_samples:
            raise ValueError(
                f"Submission file row count ({preds.shape[0]}) "
                f"does not match test set size ({n_test_samples})."
            )
    except FileNotFoundError:
        p.error(f"Submission file not found: {a.submission_file}")
    except Exception as e:
        p.error(f"Error loading submission_file: {e}")

    print("Evaluating predictions...")
    result = evaluate(preds, labels)

    print("\n--- EVALUATION RESULT ---")
    print(json.dumps(result, indent=2))


if __name__ == '__main__':
    _cli()
\end{lstlisting}

\newpage

\subsubsection{metadata.yaml}
\label{app:config.yaml}

\begin{lstlisting}[
    basicstyle=\small\ttfamily\color{monokai-fg},
    backgroundcolor=\color{monokai-bg},
    frame=single,
    rulecolor=\color{monokai-bg},
    breaklines=true,
    breakatwhitespace=true,
    numbers=left,
    numberstyle=\tiny\color{monokai-comment},
    stepnumber=1,
    xleftmargin=1.5em,
    framexleftmargin=1em,
    columns=flexible,
    keepspaces=true,
    showstringspaces=false,
    tabsize=2,
    texcl=false,
    mathescape=false,
    escapechar=\@,
    upquote=true,
    aboveskip=0.5em,
    belowskip=0.5em,
    lineskip=-0.5pt,
    stringstyle=\color{monokai-string},
    commentstyle=\color{monokai-comment}\itshape,
    identifierstyle=\color{monokai-class}
]
metric_lower_is_better: false
file_export_globs:
  - submission.csv
container_python_requirements:
  - datasets==4.0.0
evaluate_container_python_requirements:
  - datasets==4.0.0
logging_info:
  name: MathQuestionAnsweringSVAMPAccuracy
  category: Math
  research_problem: Math Question Answering
  output_type: text-generation
  dataset: ChilleD/SVAMP
  metric: Accuracy
  input_columns:
    - question_concat
  scoring_column: Answer
  shape: 300,1
  config: default
  train_split: train
  test_split: test
  custom_gold_labels: false
  custom_rad_class: false
  sota:
    - sota_paper_title: 'Achieving >97% on GSM8K: Deeply Understanding the Problems
        Makes LLMs Better Solvers for Math Word Problems'
      sota_paper_url: https://arxiv.org/pdf/2404.14963v5
      sota_score: 0.942
      sota_notes: DUP is a prompting template. Result provided is for GPT-4 with the
        GUP prompting template.
      sota_year: 2026
      sota_venue: Frontiers of Computer Science
  dataset_paper_url: https://arxiv.org/abs/2103.07191
  estimated_worst_score: 0.0
  optimal_score: 1.0
\end{lstlisting}

\newpage
\section{Harness Setup}
\label{sec:harness}

\begin{table}[ht]
\centering
\small
\renewcommand{\arraystretch}{1.2}
\begin{tabular}{lcc}
\toprule
& \textbf{\airadojo} & \textbf{\mlgym} \\
\hline
\textbf{Time/number of steps limit} &
24 hours (up to $\sim$36 including evaluation time) &
1M steps / 24 hours \\
\hline
\textbf{Can exit early} &
No &
No \\
\hline

\textbf{evaluate.py file is visible} &
Yes &
Yes \\
\hline

\textbf{Test set with labels is visible} &
No &
No \\
\hline

\textbf{Validation script} &
Agent codes during run &
Agent codes during run \\
\hline

\textbf{Validation splits} &
Cross-validation (Greedy only) &
Classical split 70-30 \\
\hline

\textbf{Scaffold implemented} &
Greedy (AIDE) &
ReAct \\
\hline

\textbf{Last submission valid always?} &
No &
No \\
\hline

\textbf{Pretrained models access} &
Yes &
Yes \\
\hline

\textbf{Num of steps / nodes captured} &
Yes &
Yes \\
\hline

\textbf{All validation scores captured} &
Yes &
Yes \\
\hline

\textbf{Every solution scored on test set} &
Yes &
Only final submitted solution \\
\hline

\textbf{Time limit per solution} &
4 hours &
1 hour \\
\hline

\textbf{Dummy submission provided} &
No &
No \\
\hline

\textbf{Num GPUs} &
1 H200 / run &
1 H200 / run \\
\hline

\textbf{Internet access} &
Yes (HF\_OFFLINE=True, but agent can set it to False) &
Yes (can be turned off) \\
\hline

\textbf{How is evaluate.py provided} &
In prompt &
In shared workspace \\
\hline

\textbf{Python version} &
3.10 &
3.10 \\
\hline

\textbf{Datasets library version} &
3.5.1 (upgrade to 4.0.0) &
4.0.0 \\

\bottomrule
\end{tabular}
\caption{Resources and constraints comparison between \airadojo{} and \mlgym{}.}
\label{app:rad-mlgym-comparison}
\end{table}

\subsection{\mlgym~system prompt}
\label{app:mlgym_details}
\lstset{
  literate={-}{{\char`-}}1,
}
\begin{tcolorbox}[
    colback=md-bg,
    colframe=md-bg,
    boxrule=0pt,
    left=6pt,
    right=6pt,
    top=6pt,
    bottom=6pt,
    enhanced,
    sharp corners,
    breakable
]
\begin{lstlisting}[
    language={},            % <- avoid language-specific catcode tweaks
    basicstyle=\ttfamily\small,
    breaklines=true,
    breakatwhitespace=true,
    columns=fullflexible,
    keepspaces=true,
    showstringspaces=false
]

SETTING: You are an autonomous Machine Learning Researcher, and you're working directly in the command line with a special interface.

The special interface consists of a file editor that shows you 1000 lines of a file at a time.
In addition to typical bash commands, you can also use the following commands to help you navigate and edit files.

COMMANDS:
open:
  docstring: opens the file at the given path in the editor. If line_number is provided, the window will be move to include that line
  signature: open "<path>" [<line_number>]
  arguments:
    - path (string) [required]: the path to the file to open
    - line_number (integer) [optional]: the line number to move the window to (if not provided, the window will start at the top of the file)

goto:
  docstring: moves the window to show <line_number>
  signature: goto <line_number>
  arguments:
    - line_number (integer) [required]: the line number to move the window to

scroll_down:
  docstring: moves the window down 1000 lines
  signature: scroll_down

scroll_up:
  docstring: moves the window down 1000 lines
  signature: scroll_up

create:
  docstring: creates and opens a new file with the given name
  signature: create <filename>
  arguments:
    - filename (string) [required]: the name of the file to create

search_dir:
  docstring: searches for search_term in all files in dir. If dir is not provided, searches in the current directory
  signature: search_dir <search_term> [<dir>]
  arguments:
    - search_term (string) [required]: the term to search for
    - dir (string) [optional]: the directory to search in (if not provided, searches in the current directory)

search_file:
  docstring: searches for search_term in file. If file is not provided, searches in the current open file
  signature: search_file <search_term> [<file>]
  arguments:
    - search_term (string) [required]: the term to search for
    - file (string) [optional]: the file to search in (if not provided, searches in the current open file)

find_file:
  docstring: finds all files with the given name in dir. If dir is not provided, searches in the current directory
  signature: find_file <file_name> [<dir>]
  arguments:
    - file_name (string) [required]: the name of the file to search for
    - dir (string) [optional]: the directory to search in (if not provided, searches in the current directory)

edit:
  docstring: replaces lines <start_line> through <end_line> (inclusive) with the given text in the open file. The replacement text is terminated by a line with only end_of_edit on it. All of the <replacement text> will be entered, so make sure your indentation is formatted properly. Python files will be checked for syntax errors after the edit. If the system detects a syntax error, the edit will not be executed. Simply try to edit the file again, but make sure to read the error message and modify the edit command you issue accordingly. Issuing the same command a second time will just lead to the same error message again.
  signature: edit <start_line>:<end_line>
<replacement_text>
end_of_edit
  arguments:
    - start_line (integer) [required]: the line number to start the edit at
    - end_line (integer) [required]: the line number to end the edit at (inclusive)
    - replacement_text (string) [required]: the text to replace the current selection with

insert:
  docstring: inserts the given text after the specified line number in the open file. The text to insert is terminated by a line with only end_of_insert on it. All of the <text_to_add> will be entered, so make sure your indentation is formatted properly. Python files will be checked for syntax errors after the insertion. If the system detects a syntax error, the insertion will not be executed. Simply try to insert again, but make sure to read the error message and modify the insert command you issue accordingly.
  signature: insert <line_number>
<text_to_add>
end_of_insert
  arguments:
    - line_number (integer) [required]: the line number after which to insert the text
    - text_to_add (string) [required]: the text to insert after the specified line

submit:
  docstring: submits your current code for evaluation on the test set and allows you to continue hill climbing. The test score will be hidden to prevent overfitting, but you'll get confirmation if the submission was valid. Only submit within triple quotes (```) should be executed - nothing else. Otherwise it will lead to parsing errors
  signature: submit



Please note that THE EDIT and INSERT COMMANDS REQUIRES PROPER INDENTATION.
If you'd like to add the line '        print(x)' you must fully write that out, with all those spaces before the code! Indentation is important and code that is not indented correctly will fail and require fixing before it can be run.

RESPONSE FORMAT:
Your shell prompt is formatted as follows:
(Open file: <path>) <cwd> $

You MUST format your output using EXACTLY two fields: DISCUSSION and command.
Your output should ALWAYS include *one* DISCUSSION section and *one* command section in EXACTLY this format:

DISCUSSION
[Your reasoning, thoughts, and plan for this step. Be specific about what you're trying to accomplish.]

```
[single command to execute]
```

CRITICAL FORMATTING RULES:
1. Start with "DISCUSSION" (no quotes, no extra formatting)
2. Write your thoughts and reasoning
3. Add a single line with three backticks: ```
4. Write exactly ONE command
5. End with three backticks: ```
6. NO text after the closing backticks
7. NO multiple commands in one response
8. NO multiple code blocks

EXAMPLE OF CORRECT FORMAT:
DISCUSSION
I need to explore the current directory to understand the project structure and identify the data files.

```
ls -la
```

You should only include a *SINGLE* command in the command section and then wait for a response from the shell before continuing with more discussion and commands. Everything you include in the DISCUSSION section will be saved for future reference. Please do not include any DISCUSSION after your action.
If you'd like to issue two commands at once, PLEASE DO NOT DO THAT! Please instead first submit just the first command, and then after receiving a response you'll be able to issue the second command.
You're free to use any other bash commands you want (e.g. find, grep, cat, ls, cd) in addition to the special commands listed above.
However, the environment does NOT support interactive session commands (e.g. python, vim), so please do not invoke them.

MACHINE LEARNING WORKFLOW:
Your goal is to achieve the best possible score on a hidden test set. Follow this systematic approach:

1. EXPLORATION PHASE:
   - Understand the task by examining data files, baseline scripts, and evaluation metrics
   - Identify the problem type (classification, regression, etc.)
   - Analyze data structure, features, and target variables

2. VALIDATION SETUP:
   - Create a proper train/validation split from the training data
   - If a separate validation set already exists, use it instead of creating your own
   - Set up evaluation code to measure performance on your validation set
   - Your validation code should be separate from evaluate.py provided but it can borrow motivation from how you'll be evaluated with evaluate.py when you do your final submission

3. BASELINE IMPLEMENTATION:
   - Start with the provided baseline script if available
   - Understand the baseline approach and its performance
   - Ensure you can reproduce baseline results

4. ITERATIVE IMPROVEMENT (MOST IMPORTANT PHASE):
   - Make incremental improvements to your model/approach
   - After each change, retrain your model and evaluate on validation set
   - Keep track of what works and what doesn't
   - Try different approaches: feature engineering, model architectures, hyperparameters
   - CONTINUE EXPERIMENTING even after finding a working solution - this is just the beginning!
   - Use hillclimbing: always try to improve your current best solution
   - Track your best validation score and keep trying to beat it

5. SUBMISSION AND CONTINUED HILL CLIMBING:
   - The `submit` command now allows you to test your solution without ending the session!
   - When you submit, your code is evaluated on the hidden test set, but the actual score is hidden to prevent overfitting
   - You'll receive confirmation that your submission was valid, then you can continue improving
   - **STRATEGY**: Submit whenever you have a working solution, then keep hill climbing to improve it further
   - Use multiple submissions throughout your session to test different approaches
   - **CONTINUE AFTER SUBMIT**: After each submit, keep experimenting with new architectures, features, and techniques
   - The goal is to maximize your compute time for hill climbing rather than just finding one working solution

IMPORTANT TIPS:
1. Always start by trying to understand the baseline script if available. This will give you an idea of one possible solution for the task and the baseline scores that you have to beat.

2. If you run a command and it doesn't work, try running a different command. A command that did not work once will not work the second time unless you modify it!

3. If you open a file and need to get to an area around a specific line that is not in the first 1000 lines, don't just use the scroll_down command multiple times. Instead, use the goto <line_number> command. It's much quicker.

4. Always make sure to look at the currently open file and the current working directory (which appears right after the currently open file). The currently open file might be in a different directory than the working directory! Note that some commands, such as 'create', open files, so they might change the current  open file.

5. When editing files, it is easy to accidentally specify a wrong line number or to write code with incorrect indentation. Always check the code after you issue an edit to make sure that it reflects what you wanted to accomplish. If it didn't, issue another command to fix it.

6. You have a limited number of actions/steps you can take in the environment. The current step and remaining number of steps will be given after every action. Use the remaining steps wisely by continuously hill climbing and making multiple submissions to test your improvements!

7. Your each action should take less than 3600 seconds to complete. If your action doesn't finish within the time limit, it will be interrupted.

8. VALIDATION IS KEY: Always use your own validation set (create or use if one exists in data directory) to evaluate model performance. Use the `submit` command to test your submission.csv with the hidden test set whenever you have a working solution, then continue improving. Note that test scores will be hidden to prevent overfitting, but your submissions will be logged.



We're currently solving the following task. Here's the task description:

TASK DESCRIPTION:
<Overview>
<Dataset Structure>
<Evaluation - evaluate.py details>


INSTRUCTIONS:
Now, you're going to train a model to improve performance on this task. Your terminal session has started and you're in the workspace root directory. You can use any bash commands or the special interface to help you. Edit all the file you need or create a new training script.
Remember, YOU CAN ONLY ENTER ONE COMMAND AT A TIME. You should always wait for feedback after every command.
When you're satisfied with all of the changes you have made, you can run your code. Your code should produce a valid submission artefact. Please follow the instructions in SUBMISSION FORMAT section above and adhere to the guidelines provided for generating submission artefacts. You can also look at the `evaluate.py` provided to you to see if you are following the correct format and naming scheme for your submission artefacts.

IMPORTANT SECURITY NOTE ABOUT TEST DATA:
  - You may see references to a `test_with_labels/` directory in evaluation scripts or documentation
  - This directory is INTENTIONALLY BLOCKED from your workspace for security reasons
  - If you try to access `data/test_with_labels/`, you will find it empty or inaccessible.
  - When you submit your solution, evaluation runs in a SEPARATE secure container that has access to the actual test labels
  - DO NOT attempt to access test_with_labels - focus on using `data/train/` and `data/test/` (without labels) for model development
  - Use validation data or cross-validation on the training set to evaluate your model during development
  - The `submit` command will run your code in the secure evaluation environment with actual test labels

  OTHER REMARKS
  - Plenty of Huggingface models are pre-downloaded to cache. You can use `from transformers.utils import hub; print(hub.default_cache_path)` to get the path of cache. If you firmly believe that the issue is not having the requested model in the cache, then you should be able to also download one yourself.
  - Be aware of the running time of the code you write, it should complete within 1 hour.
  - You are provided with a single GPU which you can use for training the models. Take its memory into consideration when proposing the solution(s).
  - All the provided input data is stored in "./data" directory

Note however that you cannot use any interactive session commands (e.g. python, vim) in this environment, but you can write scripts and run them. E.g. you can write a python script and then run it with `python <script_name>.py`.

NOTE ABOUT THE EDIT AND INSERT COMMANDs: Indentation really matters! When editing a file, make sure to insert appropriate indentation before each line!
\end{lstlisting}
\end{tcolorbox}

\subsection{\airadojo~system prompt}

\label{app:airadojo_details}
The \airadojo\ scaffold leverages the following predefined operator set: \opdraft\ initializes the search process by generating an initial population of candidate solutions. \opdebug\ attempts to identify and correct errors in buggy solutions. \opimprove\ refines valid artifacts to enhance their performance according to the evaluation criteria. While not explicitly defined as an operator, \opanalyze\ is also used to evaluate the execution output of a generated and executed solution to detect bugs and summarize empirical findings from the results. Detailed below are the prompts for each operator.

\subsubsection{Draft}
\begin{tcolorbox}[
    colback=md-bg,
    colframe=md-bg,
    boxrule=0pt,
    left=6pt,
    right=6pt,
    top=6pt,
    bottom=6pt,
    enhanced,
    sharp corners,
    breakable
]
\begin{lstlisting}[
    language={},
    basicstyle=\ttfamily\small,
    breaklines=true,
    breakatwhitespace=true,
    columns=fullflexible,
    keepspaces=true,
    showstringspaces=false
]
# Introduction:
You are a Kaggle grandmaster attending a competition.
In order to win this competition, you need to come up with an excellent and creative plan
for a solution and then implement this solution in Python. We will now provide a description of the task.

# Task Description:
{{task_desc}}

{% if memory %}
# Memory:
{{memory}}
{% endif %}

# Instructions:
## Response Format:
Your response should be a brief outline/sketch of your proposed solution in natural language (3-5 sentences),
followed by a single markdown code block (wrapped in ```) which implements this solution and prints out the evaluation metric.
There should be no additional headings or text in your response. Just natural language text followed by a newline and then the markdown code block.

## Solution sketch guideline:
This first solution design should be relatively simple, without ensembling or hyper-parameter optimization.
Take the Memory section into consideration when proposing the design,
don't propose the same modelling solution but keep the evaluation the same.
The solution sketch should be 3-5 sentences.
Propose an evaluation metric that is reasonable for this task.
Don't suggest to do EDA.
The data is already prepared and available in the `./data` directory. There is no need to unzip any files.

## Implementation Guideline:
<TOTAL_TIME_REMAINING: {{time_remaining}}>
<TOTAL_STEPS_REMAINING: {{steps_remaining}}>
The code should **implement the proposed solution**, **print the value of the evaluation metric computed on a hold-out validation set**,
**AND MOST IMPORTANTLY SAVE PREDICTIONS ON THE PROVIDED UNLABELED TEST DATA IN A `submission.csv` FILE IN THE CURRENT DIRECTORY.**
The code should be a single-file python program that is self-contained and can be executed as-is.
No parts of the code should be skipped, don't terminate the before finishing the script.
Your response should only contain a single code block.
Be aware of the running time of the code, it should complete within {{execution_timeout}}.
All the provided input data is stored in "./data" directory.
**If there is test data provided for this task, please save the test predictions in a `submission.csv` file in the "./" directory as described in the task description** This is extremely important since this file is used for grading/evaluation. DO NOT FORGET THE submission.csv file!
You can also use the current directory to store any temporary files that your code needs to create.
REMEMBER THE ./submission.csv FILE!!!!! The correct directory is important too.
The evaluation should be based on 5-fold cross-validation but only if that's an appropriate evaluation for the task at hand.

## Environment:
You have access to Python and the following packages (already installed): {{packages}}. Feel free to use additional libraries that fit the problem.

# Data Overview:
{{data_overview}}

{% if other_remarks %}
# Other Remarks:
{{other_remarks}}
{% endif %}
\end{lstlisting}
\end{tcolorbox}

\subsubsection{Debug}
\begin{tcolorbox}[
    colback=md-bg,
    colframe=md-bg,
    boxrule=0pt,
    left=6pt,
    right=6pt,
    top=6pt,
    bottom=6pt,
    enhanced,
    sharp corners,
    breakable
]
\begin{lstlisting}[
    basicstyle=\ttfamily\small,
    breaklines=true,
    breakatwhitespace=true,
    columns=fullflexible,
    keepspaces=true,
    showstringspaces=false
]
# Introduction:
You are a Kaggle grandmaster attending a competition.
Your previous solution had a bug and/or did not produce a submission.csv, so based on the information below, you should revise it in order to fix this.
Your response should be an implementation outline in natural language, followed by a single markdown code block which implements the bugfix/solution.

# Task Description:
{{task_desc}}

{% if memory %}
# Previous debugging attempts:
{{memory}}
{% endif %}

# Previous (buggy) implementation:
{{prev_buggy_code}}

# Execution output:
{{execution_output}}

# Instructions:
## Response Format:
Your response should be a brief outline/sketch of your proposed solution in natural language (3-5 sentences),
followed by a single markdown code block (wrapped in ```) which implements this solution and prints out the evaluation metric.
There should be no additional headings or text in your response. Just natural language text followed by a newline and then the markdown code block.

## Bugfix improvement sketch guideline:
You should write a brief natural language description (3-5 sentences) of how the issue in the previous implementation can be fixed.
Don't suggest to do EDA.

## Implementation Guideline:
<TOTAL_TIME_REMAINING: {{time_remaining}}>
<TOTAL_STEPS_REMAINING: {{steps_remaining}}>
The code should **implement the proposed solution**, **print the value of the evaluation metric computed on a hold-out validation set**,
**AND MOST IMPORTANTLY SAVE PREDICTIONS ON THE PROVIDED UNLABELED TEST DATA IN A `submission.csv` FILE IN THE CURRENT DIRECTORY.**
The code should be a single-file python program that is self-contained and can be executed as-is.
No parts of the code should be skipped, don't terminate the before finishing the script.
Your response should only contain a single code block.
Be aware of the running time of the code, it should complete within {{execution_timeout}}.
All the provided input data is stored in "./data" directory.
**If there is test data provided for this task, please save the test predictions in a `submission.csv` file in the "./" directory as described in the task description** This is extremely important since this file is used for grading/evaluation. DO NOT FORGET THE submission.csv file!
You can also use the current directory to store any temporary files that your code needs to create.
REMEMBER THE ./submission.csv FILE!!!!! The correct directory is important too.
The evaluation should be based on 5-fold cross-validation but only if that's an appropriate evaluation for the task at hand.

# Data Overview:
{{data_overview}}

# Other remarks
- Huggingface is set to OFFLINE mode by default. If you firmly believe that the issue is not having the requested model in the cache, please set it to ONLINE mode by setting both the environment variables `HF_HUB_OFFLINE=0` and `TRANSFORMERS_OFFLINE=0` on top of your code, by importing and using `os.environ[...] = ...`.
- Do not set/force Huggingface to OFFLINE mode as that will NOT fix any issue.
- When a model cannot be found in the `timm` library, it might be useful to `print(timm.list_models())`.
- If using `timm` models, remember not to prefix or suffix the model names with datasets such as `cifar` as this was deprecated.
\end{lstlisting}
\end{tcolorbox}

\subsubsection{Improve}
\begin{tcolorbox}[
    colback=md-bg,
    colframe=md-bg,
    boxrule=0pt,
    left=6pt,
    right=6pt,
    top=6pt,
    bottom=6pt,
    enhanced,
    sharp corners,
    breakable
]
\begin{lstlisting}[
    basicstyle=\ttfamily\small,
    breaklines=true,
    breakatwhitespace=true,
    columns=fullflexible,
    keepspaces=true,
    showstringspaces=false
]
# Introduction:
You are a Kaggle grandmaster attending a competition. You are provided with a previously developed
solution below and should improve it in order to further increase the (test time) performance.
For this you should first outline a brief plan in natural language for how the solution can be improved and
then implement this improvement in Python based on the provided previous solution.

# Task Description:
{{task_desc}}

{% if memory %}
# Memory:
{{memory}}
{% endif %}

# Previous solution:
## Code:
{{prev_code}}

# Instructions:
## Response Format:
Your response should be a brief outline/sketch of your proposed solution in natural language (3-5 sentences),
followed by a single markdown code block (wrapped in ```) which implements this solution and prints out the evaluation metric.
There should be no additional headings or text in your response. Just natural language text followed by a newline and then the markdown code block.

## Solution improvement sketch guideline:
The solution sketch should be a brief natural language description of how the previous solution can be improved.
You should be very specific and should only propose a single actionable improvement.
This improvement should be atomic so that we can experimentally evaluate the effect of the proposed change.
Take the Memory section into consideration when proposing the improvement.
The solution sketch should be 3-5 sentences.
Don't suggest to do EDA.

## Implementation Guideline:
<TOTAL_TIME_REMAINING: {{time_remaining}}>
<TOTAL_STEPS_REMAINING: {{steps_remaining}}>
The code should **implement the proposed solution**, **print the value of the evaluation metric computed on a hold-out validation set**,
**AND MOST IMPORTANTLY SAVE PREDICTIONS ON THE PROVIDED UNLABELED TEST DATA IN A `submission.csv` FILE IN THE CURRENT DIRECTORY.**
The code should be a single-file python program that is self-contained and can be executed as-is.
No parts of the code should be skipped, don't terminate the before finishing the script.
Your response should only contain a single code block.
Be aware of the running time of the code, it should complete within {{execution_timeout}}.
All the provided input data is stored in "./data" directory.
**If there is test data provided for this task, please save the test predictions in a `submission.csv` file in the "./" directory as described in the task description** This is extremely important since this file is used for grading/evaluation. DO NOT FORGET THE submission.csv file!
You can also use the current directory to store any temporary files that your code needs to create.
REMEMBER THE ./submission.csv FILE!!!!! The correct directory is important too.
The evaluation should be based on 5-fold cross-validation but only if that's an appropriate evaluation for the task at hand.

{% if other_remarks %}
# Other Remarks:
{{other_remarks}}
{% endif %}
\end{lstlisting}
\end{tcolorbox}

\subsubsection{Analyze}
\begin{tcolorbox}[
    colback=md-bg,
    colframe=md-bg,
    boxrule=0pt,
    left=6pt,
    right=6pt,
    top=6pt,
    bottom=6pt,
    enhanced,
    sharp corners,
    breakable
]
\begin{lstlisting}[
    basicstyle=\ttfamily\small,
    breaklines=true,
    breakatwhitespace=true,
    columns=fullflexible,
    keepspaces=true,
    showstringspaces=false
]
# Introduction:
You are a Kaggle grandmaster attending a competition.
You have written code to solve this task and now need to evaluate the output of the code execution.
You should determine if there were any bugs as well as report the empirical findings.

# Task Description:
{{task_desc}}

# Implementation:
{{code}}

# Execution output:
{{execution_output}}
\end{lstlisting}
\end{tcolorbox}

\newpage
\section{Compute Requirements of Benchmarks}
\label{app:compute_reqs_benchmarks}
\renewcommand{\arraystretch}{1.2} 
\begin{table}[htbp]
\centering
\small
\label{tab:benchmark-compute-summary}
\begin{NiceTabular}{p{2.8cm}X X X X X X}
\toprule

\textbf{Benchmark} & \textbf{GPU / Hardware} & \textbf{Runtime} & \textbf{Budget / Cost}
& \textbf{Notes} \\
\midrule
AIRS-Bench & 1×H200 GPUs per run & 24h/task & not specified
& 20 tasks in total, 10--20 runs/task \\
MLE-Bench & 1×A10 GPU & 24h/competition & 1,800 GPU hours total
& 75 competitions in total \\
MLGym-Bench & 0--2 GPUs per task (depending on task) & 2--4h/task & \$1/run for most LLMs, some are up to \$9
& 13 tasks in total \\
RE-Bench & 0--6 H100 GPUs (depending on task) & 8h/run & \$123/run
& 7 tasks in total; 3--5 runs/task \\
ML-Agent-Bench & not specified & 0.5--2h/task; max 5h/run & \$60 total
& 13 tasks in total \\
SWE-Bench & not specified & not specified & $\leq$ \$0.3 per task; $\sim$\$500 total
& 2294 tasks in total \\
CORE-Bench & 1×T4 GPU or CPU & 2h/task & \$4 per task; max \$6; $\leq$\$500 total
& 270 tasks in total; 3 trials/task \\
CSR-Bench & not specified & not specified & not specified
& 100 GitHub repos in total \\
Auto-Bench & not specified & not specified & \$365 total
& 6 tasks in total; 10-20 trials per task \\
SciReplicate-Bench & 1×A100 GPU & not specified & not specified
& 36 papers broken down to 100 tasks; 3 runs / task \\
PaperBench & 1×A10 GPU & Up to 12h run/paper & \$400 / paper run; \$8k total
& 20 research papers broken down to 8316 small tasks, 3 runs / paper \\
ResearchBench & not specified & not specified & not specified
& 1386 papers each with 3 tasks \\
Automated LLM Speedrunning & 8×H100 GPUs & 10h/run; max 20h/run & not specified
& 19 tasks each with 4 different levels; 3 runs per task+level \\
\bottomrule

\end{NiceTabular}
\caption{Summary of compute, runtime, and cost information for recent LLM-agent benchmarks.}

\end{table}

\newpage


\section{Cached Models}

\label{app:model_cache}
The models' cache available to our agents during the runs consists of the following $193$ pretrained models available on HuggingFace, as shown in Table \ref{tab:allowed_models}. This cache does not contain frontier models, the newest model present is deberta-v3-large released in $2021$.

\renewcommand{\arraystretch}{0.9} 

\begin{longtable}{p{.45\textwidth} p{.45\textwidth}}

\caption{HuggingFace models in the run's cache (alphabetically sorted)} \label{tab:allowed_models} \\

\toprule

\textbf{Model} & \textbf{Model} \\

\midrule

\endfirsthead

\multicolumn{2}{c}%
{{\tablename\ \thetable{} -- continued from previous page}} \\

\toprule

\textbf{Model} & \textbf{Model} \\

\midrule

\endhead

\midrule \multicolumn{2}{r}{{Continued on next page}} \\

\endfoot

\bottomrule

\endlastfoot

ai-forever--ruT5-base & ai4bharat--IndicBERTv2-MLM-only \\
ai4bharat--indic-bert & albert--albert-base-v2 \\
albert-base-v2 & albert-xxlarge-v1 \\
albert-xxlarge-v2 & allenai--longformer-base-4096 \\
allenai--scibert\_scivocab\_uncased & allenai--specter \\
anferico--bert-for-patents & BAAI--bge-large-en-v1.5 \\
BAAI--bge-small-en-v1.5 & bert-base-cased \\
bert-base-multilingual-cased & bert-base-multilingual-uncased \\
bert-base-uncased & bert-large-cased \\
bert-large-uncased & bert-large-uncased-whole-word-masking \\
bert-large-uncased-whole-word-masking-finetuned-squad & bhadresh-savani--bert-base-uncased-emotion \\
bhadresh-savani--distilbert-base-uncased-emotion & bhadresh-savani--roberta-base-emotion \\
camembert--camembert-base & camembert-base \\
cardiffnlp--twitter-roberta-base & cardiffnlp--twitter-roberta-base-emotion \\
cardiffnlp--twitter-roberta-base-sentiment & cardiffnlp--twitter-roberta-base-sentiment-latest \\
cointegrated--rut5-base & cointegrated--rut5-small \\
cross-encoder--ms-marco-MiniLM-L-6-v2 & cross-encoder--nli-deberta-v3-base \\
cross-encoder--nli-deberta-v3-large & cross-encoder--nli-roberta-base \\
cross-encoder--stsb-roberta-base & cross-encoder--stsb-roberta-large \\
deepset--roberta-base-squad2 & deepset--roberta-large-squad2 \\
deepset--xlm-roberta-base-squad2 & deepset--xlm-roberta-large-squad2 \\
DeepPavlov--rubert-base-cased & distilbert--distilbert-base-cased \\
distilbert--distilbert-base-uncased & distilbert--distilroberta-base \\
distilbert-base-cased & distilbert-base-cased-distilled-squad \\
distilbert-base-multilingual-cased & distilbert-base-uncased \\
distilbert-base-uncased-distilled-squad & distilbert-base-uncased-finetuned-sst-2-english \\
distilroberta-base & distilgpt2 \\
dmis-lab--biobert-v1.1 & facebook--bart-base \\
facebook--bart-large & facebook--bart-large-cnn \\
facebook--bart-large-mnli & facebook--fasttext-en-vectors \\
facebook--mbart-large-50 & facebook--mbart-large-cc25 \\
facebook--wav2vec2-base & facebook--wav2vec2-base-960h \\
facebook--wav2vec2-large-960h & FacebookAI--roberta-base \\
FacebookAI--roberta-large & FacebookAI--xlm-roberta-base \\
FacebookAI--xlm-roberta-large & gpt2 \\
gpt2-large & gpt2-medium \\
google--bigbird-roberta-base & google--bigbird-roberta-large \\
google--byt5-base & google--byt5-small \\
google--electra-base-discriminator & google--electra-large-discriminator \\
google--electra-large-generator & google--electra-small-discriminator \\
google--efficientnet-b0 & google--efficientnet-b3 \\
google--efficientnet-b4 & google--efficientnet-b5 \\
google--efficientnet-b6 & google--efficientnet-b7 \\
google--flan-t5-base & google--mobilebert-uncased \\
google--mt5-base & google--mt5-small \\
google--muril-base-cased & google--muril-large-cased \\
google--pegasus-xsum & google--t5-v1\_1-small \\
google--vit-base-patch16-224 & google--vit-base-patch16-224-in21k \\
google--vit-large-patch16-384 & google-bert--bert-base-cased \\
google-bert--bert-base-multilingual-cased & google-bert--bert-base-uncased \\
google-bert--bert-large-uncased & google-bert--bert-large-uncased-whole-word-masking-finetuned-squad \\
google--bert\_uncased\_L-2\_H-128\_A-2 & google--bert\_uncased\_L-4\_H-512\_A-8 \\
google-t5--t5-base & google-t5--t5-small \\
helsinki-nlp--opus-mt-de-en & Helsinki-NLP--opus-mt-en-de \\
Helsinki-NLP--opus-mt-en-es & Helsinki-NLP--opus-mt-en-fr \\
Helsinki-NLP--opus-mt-en-ROMANCE & Helsinki-NLP--opus-mt-es-en \\
Helsinki-NLP--opus-mt-fr-en & Helsinki-NLP--opus-mt-ROMANCE-en \\
Helsinki-NLP--opus-mt-ru-en & j-hartmann--emotion-english-distilroberta-base \\
j-hartmann--emotion-english-roberta-large & joeddav--distilbert-base-uncased-go-emotions-student \\
llm-blender--PairRM & microsoft--codebert-base \\
microsoft--codebert-base-mlm & microsoft--deberta-base \\
microsoft--deberta-large & microsoft--deberta-v2-xlarge \\
microsoft--deberta-v2-xxlarge & microsoft--deberta-v3-base \\
microsoft--deberta-v3-large & microsoft--deberta-v3-small \\
microsoft--DialoGPT-medium & microsoft--graphcodebert-base \\
microsoft--MiniLM-L12-H384-uncased & microsoft--mpnet-base \\
microsoft--swin-base-patch4-window7-224 & microsoft--trocr-base-printed \\
microsoft--unixcoder-base & microsoft--xtremedistil-l6-h256-uncased \\
microsoft--xtremedistil-l6-h384-uncased & openai--clip-vit-base-patch16 \\
openai--clip-vit-base-patch32 & OpenAssistant--reward-model-deberta-v3-base \\
OpenAssistant--reward-model-deberta-v3-large & OpenAssistant--reward-model-deberta-v3-large-v2 \\
prajjwal1--bert-mini & prajjwal1--bert-tiny \\
princeton-nlp--unsup-simcse-roberta-base & ProsusAI--finbert \\
roberta-base & roberta-base-openai-detector \\
roberta-large & roberta-large-mnli \\
s-nlp--roberta\_toxicity\_classifier & Salesforce--codegen-350M-mono \\
Salesforce--codet5-base & Salesforce--codet5-base-multi-sum \\
Salesforce--codet5-large & Salesforce--codet5-small \\
SamLowe--roberta-base-go\_emotions & sberbank-ai--ruT5-base \\
sberbank-ai--ruT5-large & sentence-transformers--all-distilroberta-v1 \\
sentence-transformers--all-MiniLM-L6-v2 & sentence-transformers--all-mpnet-base-v2 \\
sentence-transformers--msmarco-distilbert-base-tas-b & sentence-transformers--paraphrase-albert-small-v2 \\
sentence-transformers--paraphrase-MiniLM-L3-v2 & sentence-transformers--paraphrase-MiniLM-L6-v2 \\
sentence-transformers--paraphrase-mpnet-base-v2 & sentence-transformers--stsb-mpnet-base-v2 \\
siebert--sentiment-roberta-large-english & stanfordnlp--glove \\
t5-base & t5-large \\
t5-small & timm--efficientnet\_b4.ra2\_in1k \\
unitary--toxic-bert & unitary--unbiased-toxic-roberta \\
UrukHan--t5-russian-spell & vectara--hallucination\_evaluation\_model \\
vinai--bertweet-base & vinai--bertweet-large \\
xlnet--xlnet-base-cased & xlnet--xlnet-large-cased \\
xlnet-base-cased & xlm-roberta-base \\
xlm-roberta-large & \\

\end{longtable}

\section{Distribution of tasks SOTA venue and year}
\label{app:task_paper_info}

\begin{figure}[!htbp]
    \centering
    \begin{subfigure}{0.45\linewidth}
        \centering
        \includegraphics[width=\linewidth]{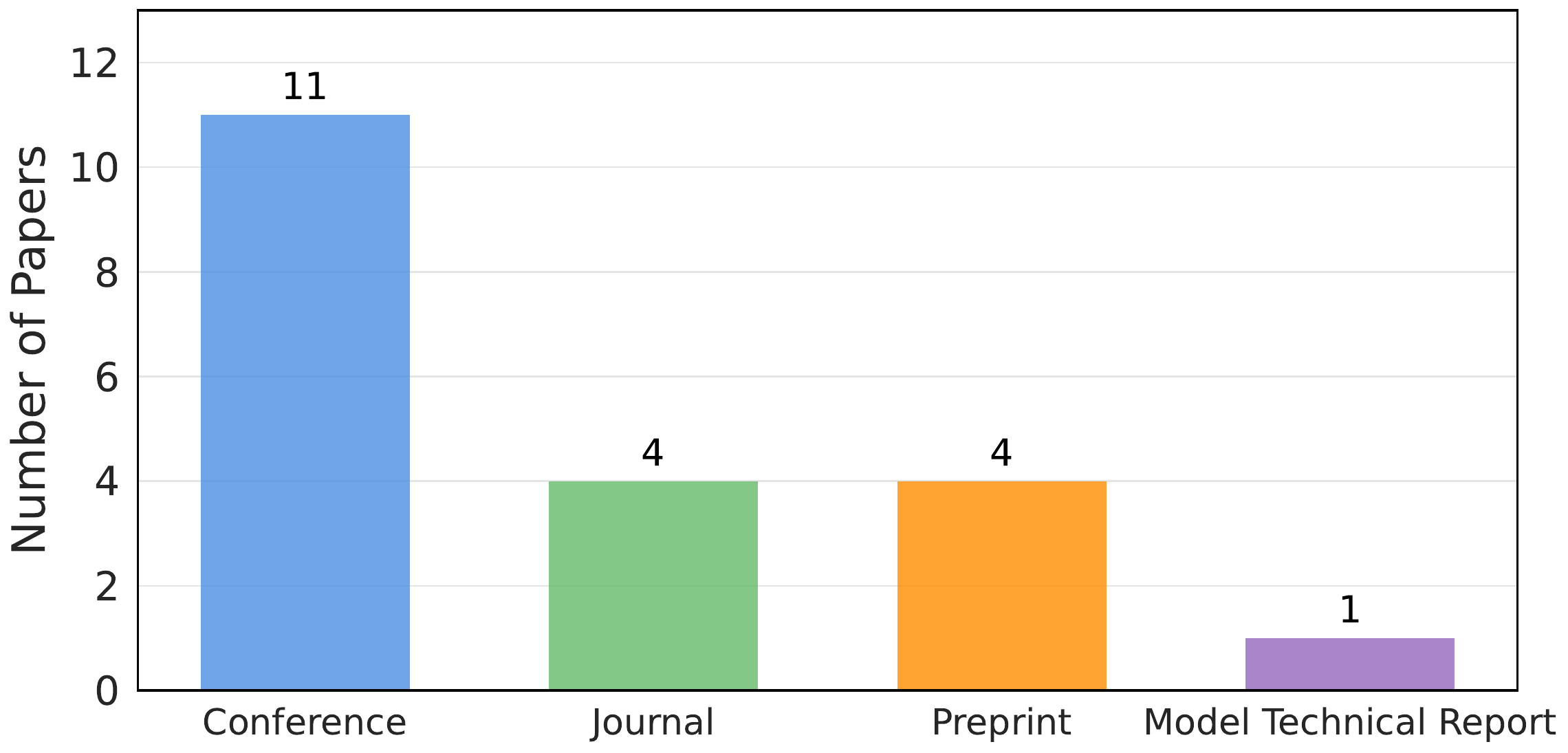}
        \vspace{-0.6cm}
        \caption{Breakdown of tasks by SOTA publication venue.}
        \label{fig:breakdown_tasks_by_venues}
    \end{subfigure}
    \hspace{2mm}
    \begin{subfigure}{0.45\linewidth}
        \centering
        \includegraphics[width=\linewidth]{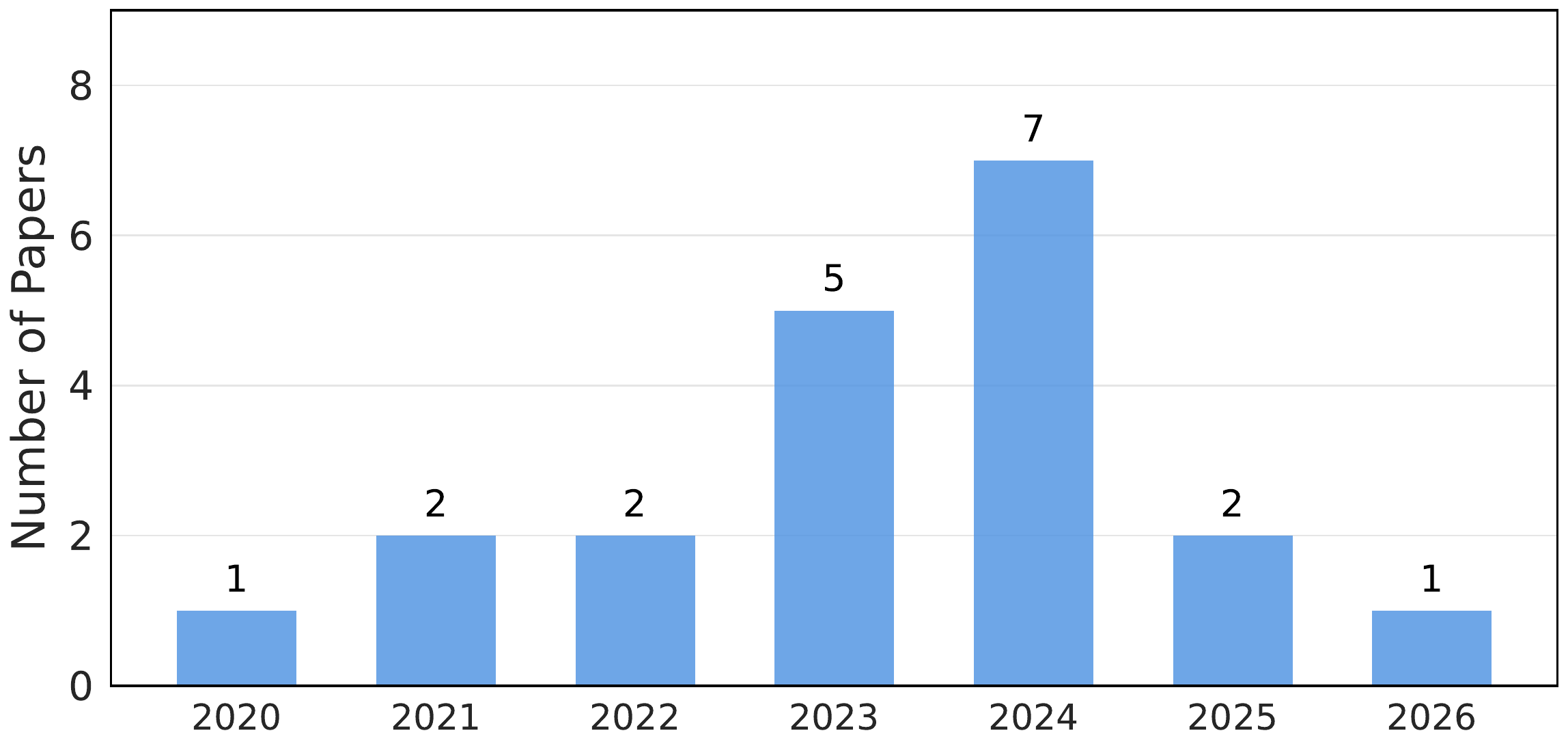}
        \vspace{-0.6cm}
        \caption{Breakdown of tasks by SOTA publication year.}
        \label{fig:breakdown_papers_by_year}
    \end{subfigure}
    \vspace{-0.25cm}
    \caption{Breakdown of tasks by SOTA publication year and venue.}
    \label{fig:venueyear}
\end{figure}

In Fig. \ref{fig:venueyear} we reprot the breakdown of \airsbench{} tasks by (a) SOTA publication venue and (b) SOTA publication year. A detailed breakdown of each venue is provided in Table \ref{tab:breakdown_paper_venue}.

\vspace{0.2cm}

\renewcommand{\arraystretch}{1.1} 
\begin{table*}[!htbp]
\centering
\rowcolors{2}{white}{faintblue}
\begin{tabularx}{\textwidth}{@{\hskip 4pt}>{\centering\arraybackslash}X@{\hskip 4pt}>{\centering\arraybackslash}p{1.2cm}@{\hskip 4pt}}
\toprule
\textbf{Venue} & \textbf{Count} \\
\midrule
ICLR & 5 \\
Preprint & 4 \\
ACL & 2 \\
AIMLSystems & 1 \\
Frontiers of Computer Science & 1 \\
Nature Communications & 1 \\
NEURIPS & 1 \\
ICML & 1 \\
EMNLP & 1 \\
Computational Linguistics & 1 \\
Model technical report & 1 \\
IEEE/ACM Transactions on Audio, Speech and Language Processing & 1 \\
\bottomrule
\end{tabularx}
\caption{Breakdown of venues where the SOTA paper was introduced.}
\label{tab:breakdown_paper_venue}
\end{table*}

\newpage

\section{AIRS-Bench Task Description}

\subsection{CodeGenerationAPPSPassAt5}

Solve coding problems by generating five distinct Python programs for each problem. It employs the APPS dataset~\citep{hendrycksapps2021}, which consists of thousands of real-world coding challenges collected from online platforms, each accompanied by a detailed natural-language problem statement and a starter code template. For each test problem, the problem statement and starter code are provided. Each program is evaluated against a set of hidden test cases, and a prediction is considered correct if at least one of the five submitted programs passes all official test cases. Model performance is assessed using the Pass@5 metric, which measures the fraction of problems solved by at least one of the five attempts.

\subsection{CodeRetrievalCodeXGlueMRR}

Retrieve relevant code snippets given natural language queries. It uses the CodeXGlue Code Search Adv dataset~\citep{lu2021codexglue}, which consists of a large corpus of code functions in Java and a set of queries describing desired functionality in natural language. For each query, the task is to search the corpus and rank code snippets by relevance, aiming to identify the correct code that implements the described functionality. During training and validation, queries are paired with the correct code snippet, while in the test set, only the queries and the code corpus are provided. Model performance is assessed using the Mean Reciprocal Rank (MRR), which measures how highly the correct code is ranked for each query.

\subsection{CoreferenceResolutionSuperGLUEWSCAccuracy}

Predict whether a pronoun refers or not to something mentioned earlier in the sentence. It uses the SuperGLUE WSC dataset~\citep{wang2019superglue}, which provides sentences with an ambiguous pronoun and a highlighted possible reference. For each example, the agent is given the sentence, the pronoun, and the possible reference. The goal is to predict whether the pronoun refers to that reference (binary classification). Model performance is measured using accuracy, which is the percentage of examples where the correct prediction is made.

\subsection{CoreferenceResolutionWinograndeAccuracy}

Identify which of two possible options a pronoun in a sentence refers to. It uses the Winogrande dataset~\citep{sakaguchi2021winogrande}, which contains sentences with an ambiguous pronoun and two possible answers. For each sentence, select the option that best fills the gap using commonsense reasoning. Model performance is assessed using accuracy, which is the percentage of correct answers.

\subsection{CvMolecularPropertyPredictionQm9MeanAbsoluteError}

Estimate a molecular property, the heat capacity at constant volume ($C_v$), from the geometry and atomic composition of a molecule. The agent is required to perform regression based on the 3D coordinates of each atom and its corresponding element. The task utilizes the QM9 dataset~\citep{ramakrishnan2014quantum}, a classic benchmark for molecular property prediction, which spans more than 10 different molecular properties determined using ab-initio density functional theory. Model performance is assessed using the mean absolute error (MAE) between the predicted and ground-truth $C_v$ values.

\subsection{GMolecularPropertyPredictionQm9MeanAbsoluteError}

Estimate a molecular property, the Gibbs free energy at $298.15$$K$  ($G$), from the geometry and atomic composition of a molecule. The agent is required to perform regression based on the 3D coordinates of each atom and its corresponding element. The task utilizes the QM9 dataset~\citep{ramakrishnan2014quantum}. Model performance is assessed using the mean absolute error (MAE) between the predicted and ground-truth $G$ values.

\subsection{GraphRegressionZincMae}

Estimate a molecular property, the constrained solubility of a molecule, from its graph structure. The task utilizes the ZINC dataset \citep{sterling2015zinc}, a widely used benchmark for graph-based molecular property prediction, which contains thousands of molecular graphs with associated solubility values. The agent is required to perform regression based on the molecular graph, where each molecule is represented as a graph with node features (atom attributes), edge indices (connectivity), and edge attributes (bond types).  Model performance is assessed using the mean absolute error (MAE) between the predicted and ground-truth solubility values.

\subsection{MathQuestionAnsweringSVAMPAccuracy}

Solve math word problems by reading a short story and answering a specific numerical question. The agent is required to predict the correct numerical answer based on the provided narrative and question, which may involve operations such as addition, subtraction, multiplication, or division. The task utilizes the SVAMP dataset \citep{patel2021nlp}, a benchmark for evaluating mathematical reasoning and problem-solving abilities in natural language. Each example consists of a description, a question, and the correct answer. Model performance is assessed using accuracy, which is the percentage of examples where the predicted answer matches ground-truth.

\subsection{QuestionAnsweringDuoRCAccuracy}

Answer questions based on a large context from movie plots. For each example, the agent is provided with the title of a story, a detailed plot summary, and a question about the story. The task is to determine whether the answer to the question is present in the context, and if so, to select the correct answer from a list of candidate answers. The DuoRC dataset \citep{saha2018duorc} is used for this task, which contains diverse and challenging reading comprehension questions requiring reasoning over long narrative texts. Model performance is assessed using accuracy, which measures the percentage of questions for which the agent correctly identifies whether an answer exists and, if so, selects the exact answer from the provided candidates.

\subsection{QuestionAnsweringEli5Rouge1}

Answer open-ended questions using long-form, explanatory responses. For each example, the agent is provided with a question, a detailed context, and is required to generate a comprehensive, human-readable answer. The task utilizes the the ELI5 (Explain Like I'm Five) dataset, containing questions and high-quality, crowd-sourced answers \citep{fan2019eli5}. Model performance is assessed using the ROUGE-1 F-measure, which evaluates the overlap of unigrams (words) between the generated answer and the reference answer, measuring the quality and relevance of the response.

\subsection{QuestionAnsweringFinqaAccuracy}

Answer financial reasoning questions based on a combination of textual context and tabular data. For each example, the agent is provided with a question, supporting context, and a table containing relevant financial information. The task utilizes the FinQA dataset \citep{chen2021finqa}, which is designed to evaluate complex question answering and reasoning over both natural language and structured tables in the financial domain. Model performance is assessed using accuracy, which measures the percentage of questions for which the predicted answer exactly matches the ground-truth, accounting for both numerical and textual equivalence.

\subsection{ReadingComprehensionSquadExactMatch}

Extract answers to questions from context paragraphs in a reading comprehension setting. For each example, the agent is provided with a title, a context paragraph, and a question about the context. The task is to extract a span of text from the context that answers the question. The dataset uses the SQuAD dataset \citep{rajpurkar2016squad}, which is a widely adopted benchmark for machine reading comprehension. Model performance is assessed using the Exact Match metric, which measures the percentage of predictions that exactly match one of the ground-truth answers provided in the dataset.

\subsection{R2AbsMolecularPropertyPredictionQm9MeanAbsoluteError}

Estimate a molecular property, the electronic spatial extent ($R^2$), from the geometry and atomic composition of a molecule. The agent is required to perform regression based on the 3D coordinates of each atom and its corresponding element. The task utilizes the QM9 dataset~\citep{ramakrishnan2014quantum}. Model performance is assessed using the mean absolute error (MAE) between the predicted and ground-truth $R^2$ values.

\subsection{SentimentAnalysisYelpReviewFullAccuracy}

Perform sentiment analysis on user-generated reviews from Yelp. For each example, the agent is provided with the text of a Yelp review and is required to predict the corresponding sentiment label, which represents the star rating assigned by the user. The dataset employed derives from the Yelp Dataset Challenge 2015 \cite{asghar2016yelp}, containing reviews labeled as one of five classes: `1 star', `2 stars', `3 stars', `4 stars', or `5 stars' (encoded as 0, 1, 2, 3, or 4). Model performance is assessed using accuracy, which measures the percentage of reviews for which the predicted label exactly matches the ground-truth rating.

\subsection{TextualClassificationSickAccuracy}

Classify the entailment relationship between two sentences. For each example, the agent is provided with a pair of sentences A and B, and must predict whether the relationship is:
\textit{entailment}, i.e. sentence B can be logically inferred from sentence A; \textit{neutral},  there is no clear logical relationship; \textit{contradiction}, sentence B contradicts sentence A. The task uses the SICK dataset \cite{marelli-etal-2014-sick}, a standard benchmark for evaluating models on sentence-level semantic relatedness. Model performance is assessed using accuracy, which measures the percentage of predictions that exactly match the ground-truth label.

\subsection{TextualSimilaritySickSpearmanCorrelation}

Estimate the semantic relatedness between two sentences by predicting a similarity score from 0 (completely unrelated) to 5 (highly related). For each example, the agent is provided with a pair of sentences, and must output a floating-point score reflecting their degree of semantic similarity. The task uses the SICK dataset \cite{marelli-etal-2014-sick}. Model performance is assessed using the Spearman correlation coefficient between the predicted scores and the ground-truth scores, measuring how well the model's ranking of sentence pairs matches the human-annotated rankings.

\subsection{TimeSeriesForecastingKaggleWebTrafficMASE}

Perform time series forecasting over the Kaggle Web Traffic dataset, which is part of the Monash Time Series Forecasting Repository. The repository is an extensive collection of time series datasets curated by Monash University and a widely adopted benchmark in the field \citep{godahewa2021monash}. The dataset contains 145063 daily time series representing the number of hits or web traffic for a set of Wikipedia pages from 01/07/2015 to 10/09/2017 used by the Kaggle web traffic forecasting competition \citep{maggie2017webtraffic}. The goal of the task is to predict the future trajectory of the series by forecasting 59 time steps ahead. Model performance is assessed using the mean absolute scaled error (MASE)  between the predicted and ground-truth values in the time series.

\subsection{TimeSeriesForecastingRideshareMAE}

Perform time series forecasting over the Rideshare dataset, which is part of the Monash Time Series Forecasting Repository \citep{godahewa2021monash}. The dataset contains hourly time series representations of attributes related to Uber and Lyft rideshare services for various locations in New York between 26/11/2018 and 18/12/2018. The dataset contains 2304 individual time series, each capturing different aspects of rideshare demand and pricing, including pickup requests, pricing variations, and service availability across different geographic zones and time periods. The goal of the task is to predict the future trajectory of the series by forecasting 48 time steps ahead. Model performance is assessed using the mean absolute error (MAE)  between the predicted and ground-truth values in the time series.

\subsection{TimeSeriesForecastingSolarWeeklyMAE}

Perform time series forecasting over the Rideshare dataset, which is part of the Monash Time Series Forecasting Repository \citep{godahewa2021monash}. The dataset provides weekly aggregated solar power generation and forecast data for a large set of simulated photovoltaic (PV) plants across the United States. The dataset captures the dynamics of solar power generation, including seasonal variations, weather-dependent fluctuations, and geographic diversity across different climate zones. The goal of the task is to predict the future trajectory of the series by forecasting 5 time steps ahead. Model performance is assessed using the mean absolute error (MAE)  between the predicted and ground-truth values in the time series.

\subsection{U0MolecularPropertyPredictionQm9MeanAbsoluteError}

Estimate a molecular property, the atomization energy at 0 K ($U_0$), from the geometry and atomic composition of a molecule. The agent is required to perform regression based on the 3D coordinates of each atom and its corresponding element. The task utilizes the QM9 dataset~\citep{ramakrishnan2014quantum}. Model performance is assessed using the mean absolute error (MAE) between the predicted and ground-truth $U_0$ values.

\end{document}